\documentclass[journal]{IEEEtran}
\usepackage{graphicx}
\usepackage{amsmath,amssymb} 
\usepackage{color}
\usepackage{color}
\usepackage{epsfig}
\usepackage{listings}
\usepackage{framed}
\usepackage{caption,subfig}
\usepackage{mdwlist}
\usepackage[lined, boxed]{algorithm2e}
\usepackage{multirow}
\usepackage{caption}
\usepackage{amsthm}

\usepackage{hyperref}

\newtheorem{theorem}{Theorem}

\newtheorem{proposition}[theorem]{Proposition}

\begin{document}

\title{Compositional Dictionaries for Domain Adaptive Face Recognition}

\author{Qiang Qiu, and Rama~Chellappa,~\IEEEmembership{Fellow,~IEEE}
\thanks{Q.~Qiu is at the Department of Electrical and Computer Engineering, Duke University, Durham, NC 27708 USA. R.~Chellappa is with the Center for
Automation Research, UMIACS, University of Maryland, College Park,
MD 20742 USA {(e-mail: qiang.qiu@duke.edu, rama@umiacs.umd.edu})}.
}

\maketitle

\begin{abstract}
We present a dictionary learning approach to compensate for the transformation of faces due to changes in view point, illumination, resolution, etc. The key idea of our approach is to force domain-invariant sparse coding, i.e., design a consistent sparse representation of the same face in different domains. In this way, classifiers trained on the sparse codes in the source domain consisting of frontal faces  can be applied to the target domain (consisting of faces in different poses, illumination conditions, etc) without much loss in recognition accuracy.
The approach is to first learn a domain base dictionary, and then describe each domain  shift (identity, pose, illumination) using a sparse representation over the base dictionary.
The dictionary adapted to each domain is expressed as sparse linear combinations of the base dictionary.
In the context of face recognition, with the proposed compositional dictionary approach, a face image can be decomposed into sparse representations for a given subject, pose and illumination respectively.
This approach has three advantages: first, the extracted sparse representation for a subject is consistent across domains and enables pose and illumination insensitive face recognition.
Second, sparse representations for pose and illumination can subsequently be used to estimate the pose and illumination condition of a face image.
Finally, by composing sparse representations for subject and the different domains, we can also perform pose alignment and illumination normalization.
Extensive experiments using two public face datasets are presented to demonstrate the effectiveness of the proposed approach for face recognition.
\end{abstract}


\section{Introduction}

\begin{figure} [t]
\centering
\includegraphics[angle=0, height=0.23\textwidth, width=.5\textwidth]{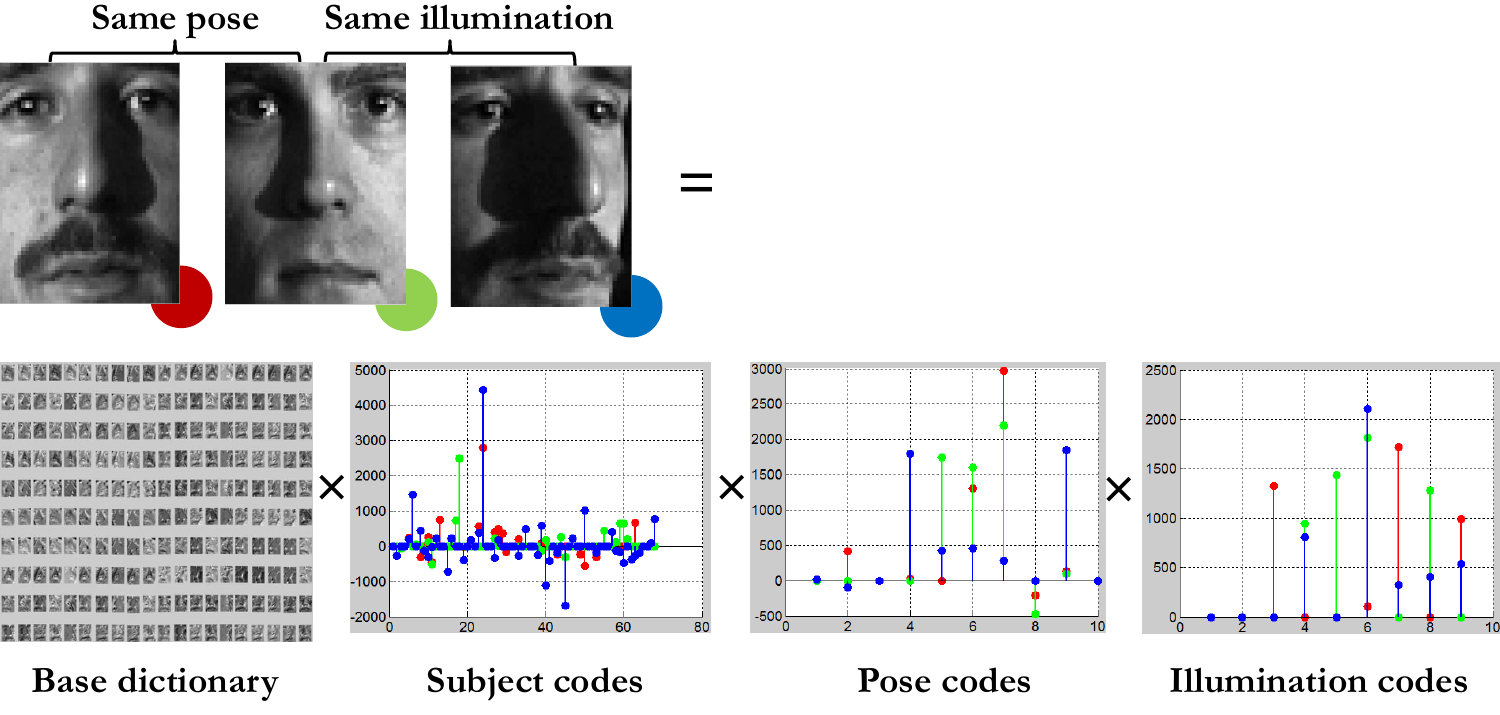}
\caption{Trilinear sparse decomposition. Given a domain base dictionary, an unknown face image is decomposed into sparse representations for each subject, pose and illumination respectively. The domain-invariant subject (sparse) codes are used for pose and illumination insensitive face recognition. The pose and illumination codes are also used to estimate the pose and lighting condition of a given face. Composing subject codes with corresponding domain codes enables pose alignment and illumination normalization.
Note that the proposed domain-invariant sparse coding assigns similar subject codes to the first and third faces (red and blue), similar pose codes to the first and second faces (red and green), and similar illumination codes to the second and third faces (green and blue).
}
\label{fig:trisparsity}
\end{figure}

Many image recognition algorithms  often fail while experiencing a significant visual domain shift, as they expect the test data to share the same underlying distribution as the training data.
A visual domain shift is common and natural in the context of face recognition. Such domain shift is due to changes in poses, illumination, resolution, etc.. Domain adaptation \cite{adap-theory} is a promising methodology for handling the domain shift by utilizing knowledge in the source domain for problems in a different but related target domain.
\cite{Daume} is one of the earliest works on semi-supervised domain adaptation, where they model data with three underlying distributions: source domain data distribution, target domain data distribution and a distribution of data that is common to both domains.
\cite{action-wv} follows a similar model in handling view point changes in the context of activity recognition, where they assume some activities are observed in both source and target domains, while some other activities are only in one of the domains. Under the above assumption, certain hyperplane-based features trained in the source domain are adapted to the target domain for improved classification. Domain adaptation for object recognition is studied in \cite{adapt11-raghu}, where the subspaces of the source domain, the target domain and the potential intermediate domains are modeled as points on the Grassmann manifold. The shift between domains is learned by exploiting the geometry of the underlying manifolds. A good survey on domain adaptation can be found in \cite{adapt11-raghu}.

Face recognition across domain, e.g., pose and illumination, has proved to be a challenging problem \cite{lightfield,s-smd,tensorface1}.
In \cite{lightfield}, the eigen light-field (ELF) algorithm is presented for face recognition across pose and illumination. This algorithm operates by estimating the eigen light field or the plenoptic function of the subject's head using all the pixels of various images. In \cite{s-smd,smd}, face recognition across pose is performed using stereo matching distance (SMD). The cost to match a probe image to a gallery image is used to evaluate the similarity of the two images.
For near frontal faces,  recent face alignment efforts  such as \cite{face-fern, one-milli-align,cmu-align} have been shown to be effective.
For face recognition across severe pose and/or illumination variations, ELF and SMD methods still report state-of-the-art results.
The proposed compositional dictionary learning approach shows comparable performance to these two methods for face recognition across domain  shifts due to pose and illumination variations. In addition, our approach can also be used to classify  the pose and lighting condition of a face, and perform pose alignment and illumination normalization.

The approach presented here shares some of the attributes of the Tensorfaces method proposed in \cite{tensorface1,tensorface2, tf-Savvides1}, but significantly differs in many aspects. In the Tensorfaces method, face images observed in different domains, i.e., faces imaged in different poses under different illuminations, form a face tensor. Then a multilinear analysis is performed on the face tensor using the $N$-mode SVD decomposition to obtain a core tensor and multiple mode matrices, each for a different domain aspect.
The $N$-mode SVD decomposition is similar to the proposed multilinear sparse decomposition shown in Fig.~\ref{fig:trisparsity}, where a given unknown image is decomposed into multiple sparse representations for the given subject, pose and illumination respectively.
However, we show through experiments that our method based on sparse decomposition significantly outperforms the $N$-mode SVD decomposition for face recognition across pose and illumination.
Another advantage of the proposed method approach over Tensorfaces is that, the proposed approach provides explicit sparse representations for each subject and each visual domain, which can be used for subject classification and domain estimation. Instead, Tensorfaces performs subject classification through exhaustive projections and matchings. Another work similar to Tensorfaces is discussed in \cite{separatingstyle}, where a bilinear analysis is presented for face matching across domains.
In \cite{separatingstyle}, a 2-mode SVD decomposition is  performed.

This paper makes the following main contributions:
\begin{itemize*}
 \item The proposed domain-invariant sparse coding  enables a  robust way for multilinear decomposition, and provides explicit sparse representations for out-of-training samples.  Note that tensor-based methods obtain representations for out-of-training samples through exhaustive projections and matchings.
  \item The proposed domain base dictionary learning provides a base dictionary that is independent of subjects and domains, and we express the dictionary adapted to a specific domain as sparse linear combinations of  base dictionary atoms using sparse representation of the domain under consideration.
\item  A face image is decomposed into sparse representations for subject, pose and illumination respectively. The domain-invariant subject (sparse) codes are used for pose and illumination insensitive face recognition. The pose and illumination codes are also used to estimate the pose and lighting condition of a given face. Composing subject codes with corresponding domain codes enables pose alignment and illumination normalization.
\end{itemize*}

The remainder of the paper is organized as follows:  Section~\ref{sec:background} discusses some details about sparse decomposition and multilinear image analysis.
 In Section~\ref{sec:formulation}, we formulate  the compositional dictionary
learning problem for face recognition.
  In Section~\ref{sec:dadl}, we present the proposed compositional dictionary learning approach, which consists of algorithms to learn a domain base dictionary, and
perform domain-invariant sparse coding.
Experimental evaluations are given in Section~\ref{sec:experiment} on two public
face datasets.
Finally, Section~\ref{sec:conclusion} concludes the paper.

\section{Background}
\label{sec:background}

\subsection{Sparse Decomposition}

Sparse signal representations have recently
drawn much attention in vision, signal and image processing research \cite{dict_IEEE}, \cite{YiMa_SR}, \cite{iccv11-qiu}, \cite{dadl}.
This is mainly due to the fact that signals and images
of interest can be sparse in some dictionary.  Given an over-complete dictionary $\mathbf{D}$ and a signal $\mathbf{y}$, finding a sparse representation of $\mathbf{y}$ in $\mathbf{D}$ entails solving the following optimization problem
\begin{equation}\label{eq:SR}
\mathbf{\hat{x}}=\arg\min_{\mathbf{x}}\| \mathbf{x} \|_0
\textrm{\;subject\; to\;} \mathbf{y}=\mathbf{D}\mathbf{x},
\end{equation}
where the
$\ell_{0}$ sparsity measure $\|\mathbf{x}\|_{0}$ counts the number of
nonzero elements in the vector $\mathbf{x}$.  Problem (\ref{eq:SR}) is NP-hard and cannot be solved in a polynomial time.  Hence, approximate solutions are usually sought  \cite{BP}, \cite{omp}, \cite{greedgood}.

The dictionary $\mathbf{D}$ can be either based on a mathematical model
of the data or it can be trained directly from the data \cite{Fields}. It has
been observed that learning a dictionary directly from training
rather than using a predetermined dictionary (such as wavelet
or Gabor) usually leads to better representation and hence
can provide improved results in many practical applications
such as restoration and classification \cite{dict_IEEE}, \cite{YiMa_SR}.

Various algorithms have been developed for the task of training a dictionary from examples.  One of the most commonly used algorithms is the K-SVD algorithm \cite{Elad_KSVD}.
Let $\mathbf{Y}$ be a set of $N$ input signals in a $n$-dimensional feature space  $\mathbf{Y}=[\mathbf{y}_1...\mathbf{y}_N], ~\mathbf{y}_i \in \mathbb{R}^{n}$.
In K-SVD, a dictionary with a fixed number of $K$ items is learned by finding a solution iteratively to the following problem:
\begin{eqnarray}
\mathbf{\arg \min_{D,X}\|Y-DX\|}_F^2 & ~~~s.t.~  \forall i, \|\mathbf{x}_i\|_0 \leq t
\label{eqt:representationerr}
\end{eqnarray}
where $\mathbf{D}=[\mathbf{d}_1...\mathbf{d}_K], ~\mathbf{d}_i \in \mathbb{R}^{n}$ is the learned dictionary, $\mathbf{X}=[\mathbf{x}_1,...,\mathbf{x}_N], ~\mathbf{x}_i \in \mathbb{R}^{K}$ are the sparse codes of input signals $\mathbf{Y}$, and $T$ specifies the sparsity that each signal has fewer than $t$ items in its decomposition. Each dictionary item $\mathbf{d}_i$ is $l_2$-normalized.

\subsection{Multilinear Image Analysis}
\label{sec:tensor}

Linear methods are popular in facial image analysis,  such as principal components analysis (PCA) \cite{eigenface}, independent component analysis (ICA) \cite{face-survey}, and linear discriminant analysis (LDA) \cite{fisherfaces}. These conventional linear analysis methods work best when variations in domains, such as pose and illumination, are not present.
When any visual domain is allowed to vary, the linear subspace representation above does not capture such variation well.

Under the assumption of Lambertian reflectance, Basri and Jacobs \cite{9point} showed that images of an object obtained under a wide variety of lighting conditions can be approximated accurately with a 9-dimensional linear subspace. \cite{harmonics} utilizes the fact that 2D harmonic basis images at different poses are related by close-form linear transformations \cite{reflectionTOG}, \cite{GroupTheory}, and extends the 9-dimensional  illumination linear space with additional pose information encoded in a linear transformation matrix. The success of these methods suggests the feasibility of decomposing a face image into separate representations for subject and individual domains, e.g. associated pose and illumination, through multilinear algebra.

\begin{figure}[ht]
\centering
\includegraphics[angle=0, height=0.12\textwidth, width=.25\textwidth]{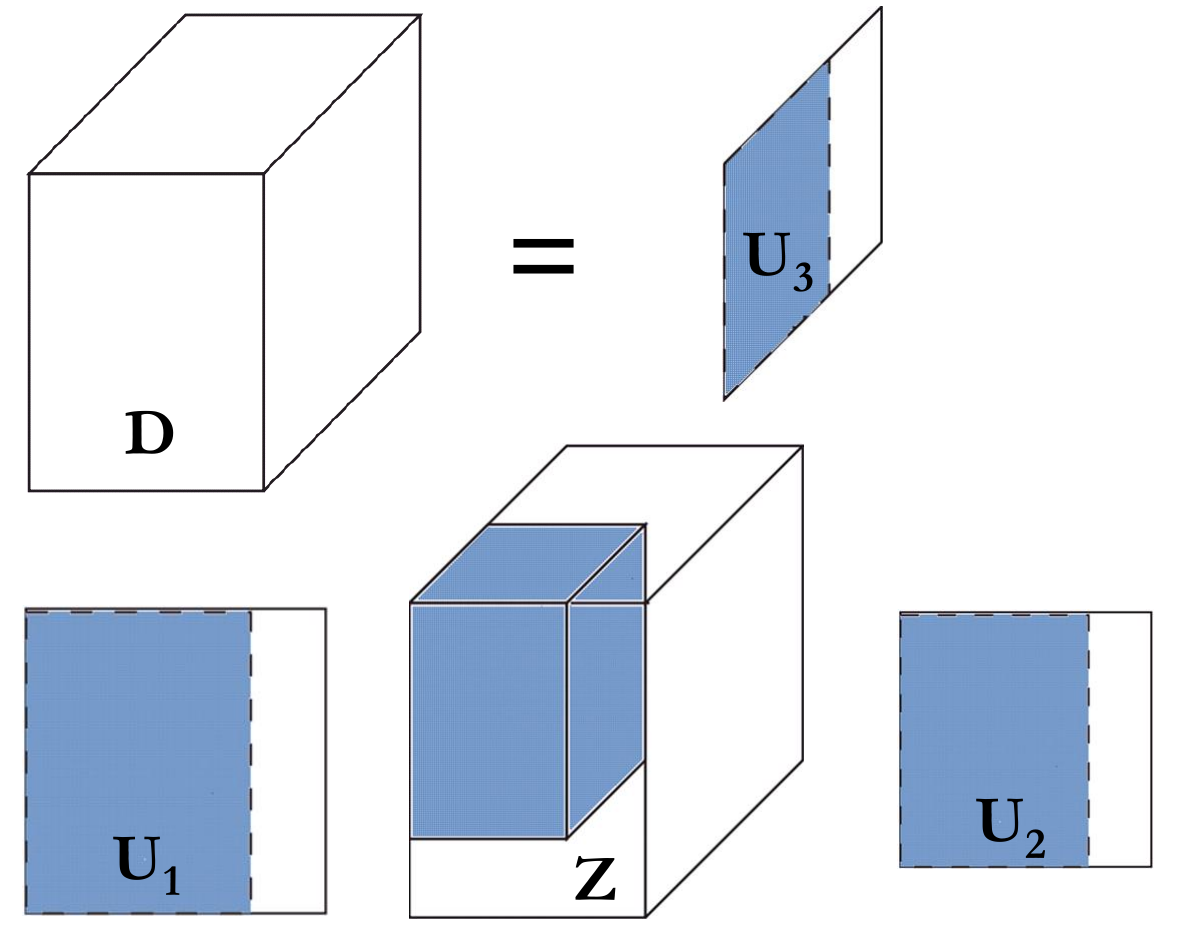}
\caption{An $N$-mode SVD (N=3 is illustrated) \cite{tensorface1}.}
 \label{fig:n-svd}
\end{figure}

A multilinear image analysis approach, called Tensorfaces, has been discussed in \cite{tensorface1}, \cite{tensorface2}, \cite{tf-Savvides1}. Tensor is a multidimensional generalization of a matrix. An $N$-th order tensor $\mathcal{D}$ is an $N$-dimensional matrix comprising $N$ spaces. $N$-mode SVD, illustrated in Fig.~\ref{fig:n-svd}, is an extension of SVD that decomposes the tensor as the product of $N$-orthogonal spaces, where Tensor $\mathcal{Z}$, the core tensor, is analogous to the diagonal singular value matrix in SVD.  The mode matrix $\mathbf{U_n}$ contains the orthonormal vectors spanning the column space of  mode-$n$ flattening of $\mathcal{D}$, i.e., the rearranged tensor elements that form a regular matrix \cite{tensorface1}.

Consider the illustrative example presented in \cite{tensorface1}. Given face images of 28 subjects, in 5 poses, 3 illuminations and 3 expressions, and each image contains 7943 pixels, we obtain a face tensor $\mathcal{D}$ of size $28 \times 5 \times 3 \times 3 \times 7943$. Suppose we apply a multilinear analysis to the face tensor $\mathcal{D}$ using the 5-mode decomposition as (\ref{eq:n-svd}).
\begin{align} \label{eq:n-svd}
\mathcal{D} = \mathcal{Z} \times \mathbf{U}_{subject} \times \mathbf{U}_{pose} \times \mathbf{U}_{illum} \times \mathbf{U}_{expre} \times \mathbf{U}_{pixels}
\end{align}
where the $28 \times 5 \times 3 \times 3 \times 7943$ core tensor $\mathcal{Z}$ governs the interaction between the factors represented in the 5 mode matrices, and each of the mode matrix $\mathbf{U}_n$ represents subjects and respective domains. For example, the $k^{th}$ row of the $28 \times 28$ mode matrix $\mathbf{U}_{subject}$ contains the coefficients for subject $k$, and the $j^{th}$ row of $5 \times 5$ mode matrix $\mathbf{U}_{pose}$ contains the coefficients for pose $j$.

Tensorfaces perform subject classification through exhaustive projections and matchings.
In the above examples, from the training data, each subject is represented with a 28-sized vector of coefficients to the $28 \times 5 \times 3 \times 3 \times 7943$ { basis} tensor in (\ref{eq:base-tensor})
\begin{align} \label{eq:base-tensor}
\mathcal{B} = \mathcal{Z} \times \mathbf{U}_{pose} \times \mathbf{U}_{illum} \times \mathbf{U}_{expre} \times \mathbf{U}_{pixels}
\end{align}
One can then obtain the basis tensor for a particular pose $j$, illumination $l$, and expression $e$ as a $28 \times 1 \times 1 \times 1 \times 7943$ sized subtensor $\mathbf{B}_{j,l,e}$.
The subject coefficients of a given unknown face image are obtained by exhaustively projecting this image into a set of candidate basis tensors for every $j,l,e$ combinations.
The resulting vector that yields the smallest distance to one of the rows in $\mathbf{U}_{subject}$ is adopted as the coefficients for the subject in the test image. In a similar way, one can obtain the coefficient vectors for pose and illumination associated with such a test image.

\section{Problem Formulation}
\label{sec:formulation}

\begin{figure}[ht]
\centering
\includegraphics[angle=0, height=0.32\textwidth, width=.5\textwidth]{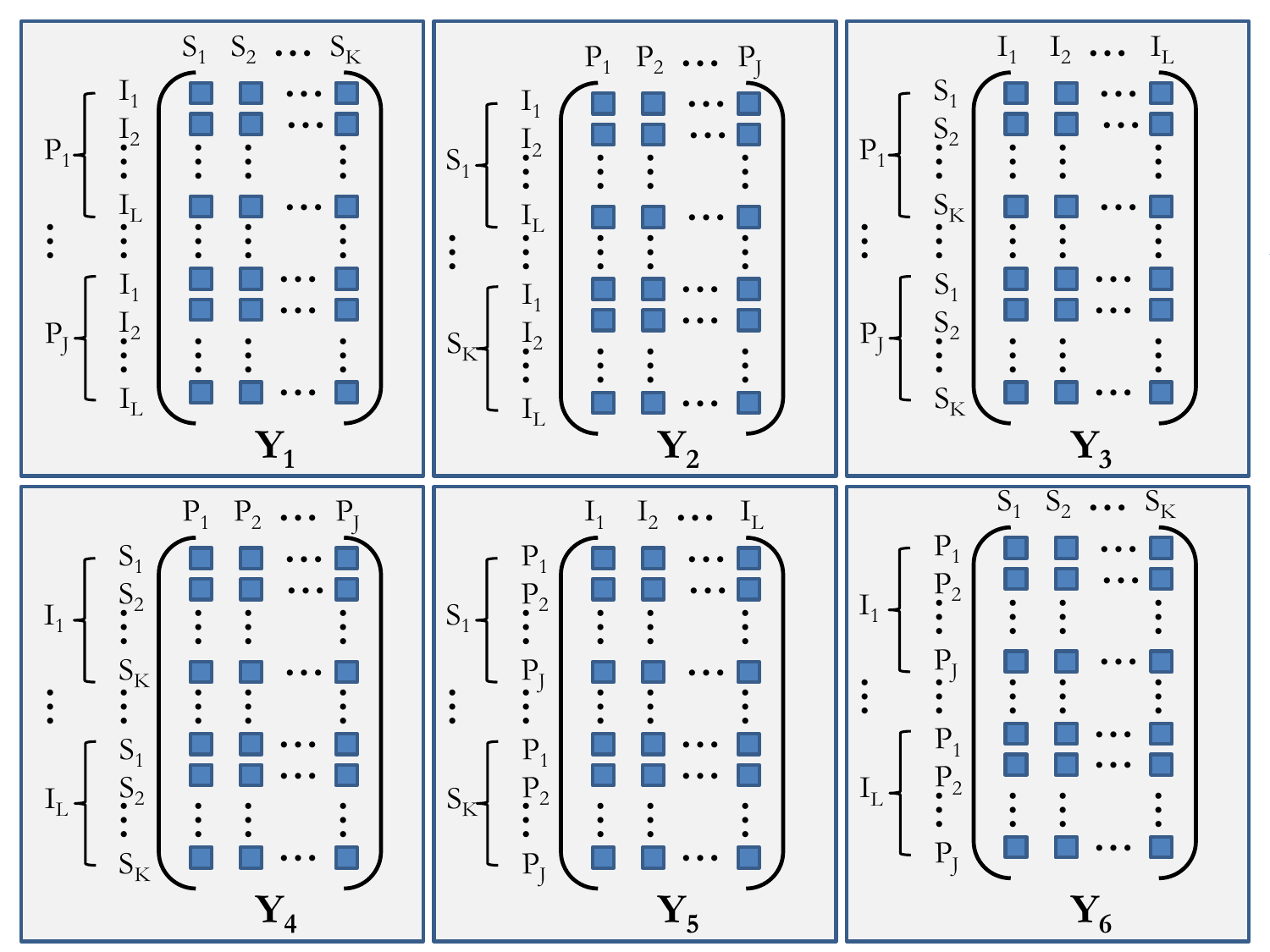}
\caption{Six forms of arranging face images of $K$ subjects in $J$ poses under $L$ illumination conditions.  Each square denotes a face image in a column vector form.}
 \label{fig:6form}
\end{figure}

In this section, we formulate the compositional dictionary learning (CDL) approach for face recognition. It is noted that our approach is general and applicable to both image and non-image data.
Let $\mathbf{Y}$ denote a set of $N$ signals (face images) in an $n$-dim feature space $\mathbf{Y}=[\mathbf{y}_1,...,\mathbf{y}_N], ~\mathbf{y}_i \in \mathbb{R}^{n}$.
Given that face images are from $K$ different subjects $[S_1, \cdots, S_K]$, in $J$ different poses $[P_1, \cdots, P_J]$, and under $L$ different illumination conditions $[I_1, \cdots, I_L]$, $\mathbf{Y}$ can be arranged in six different forms as shown in Fig.~\ref{fig:6form}. We assume here that one image is available for each subject under each pose and illumination, i.e., $N = K \times J \times L$.

\begin{figure}[ht]
\centering
\includegraphics[angle=0, height=0.1\textwidth, width=.25\textwidth]{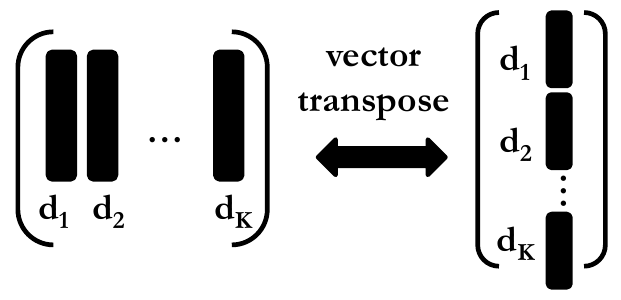}
\caption{The vector transpose operator.}
 \label{fig:vt}
\end{figure}

$\mathbf{A}$ denotes the sparse coefficient matrix of $J$ different poses, $\mathbf{A}=[\mathbf{a}_1,...,\mathbf{a}_J]$, where $\mathbf{a}_j$ is the sparse representation for the pose $P_j$. Let $dim(\mathbf{a}_j)$ denote the chosen size of sparse code vector $\mathbf{a}_j$, and $dim(\mathbf{a}_j) \le J$.
$\mathbf{B}$ denotes the sparse code matrix of $K$ different subjects, $\mathbf{B}=[\mathbf{b}_1,...,\mathbf{b}_K]$, where $\mathbf{b}_k$ is the domain-invariant sparse representation for the subject $S_k$, and $dim(\mathbf{b}_k) \le K$.
$\mathbf{C}$ denotes the sparse coefficient matrix of $L$ different illumination conditions, $\mathbf{C}=[\mathbf{c}_1,...,\mathbf{c}_L]$, where $\mathbf{c}_l$ is the sparse representation for the illumination condition $I_l$ and $dim(\mathbf{c}_l) \le L$.
The domain base dictionary $\mathbf{D}$ contains $dim(\mathbf{a}_j) \times dim(\mathbf{b}_k) \times dim(\mathbf{c}_l)$ atoms arranging in a similar way as Fig.~\ref{fig:6form}. Each dictionary atom is in
the $\mathbb{R}^{n}$ space.

 Any of the six forms in Fig.~\ref{fig:6form} can be transformed into another through a sequence of vector transpose operations.
As illustrated in Fig.~\ref{fig:vt}, a vector transpose operation is to consider (stacked) image vectors in Fig.~\ref{fig:6form} as values and perform typical matrix transpose operation.
    For simplicity, we define six aggregated vector transpose operations
     $\{T_i\}_{i=1}^6$. For example, $T_i$ transforms an input matrix, which is in any of the six forms, into the $i$-th form defined in Fig.~\ref{fig:6form} {(note that  three out of six operations are actually used)}.

Let $\mathbf{y}_k^{jl}$ be a face image of subject $S_k$ in pose $P_j$ under illumination $I_l$.
The dictionary adapted to pose $P_j$ and illumination $I_l$ is expressed as
\[
[[\mathbf{D}^{T_2}\mathbf{a}_j]^{T_3}\mathbf{c}_l]^{T_1}.
\]
$\mathbf{y}_k^{jl}$  can be sparsely represented using this dictionary as,
\[
\mathbf{y}_k^{jl} = [[\mathbf{D}^{T_2}\mathbf{a}_j]^{T_3}\mathbf{c}_l]^{T_1}\mathbf{b}_k,
\]
where the subject sparse codes $\mathbf{b_k}$ are independent of both $P_j$ and $I_l$.  In this way, we can represent Fig.~\ref{fig:6form} in a compact matrix form as shown in (\ref{eqn:6form-eq}).

\begin{subequations} \label{eqn:6form-eq}
\begin{align}
\mathbf{Y}_1 =[[\mathbf{D}^{T_3}\mathbf{C}_1]^{T_2}\mathbf{A}_1]^{T_1}\mathbf{B}_1 \label{form1}\\
\mathbf{Y}_2 =[[\mathbf{D}^{T_3}\mathbf{C}_2]^{T_1}\mathbf{B}_2]^{T_2}\mathbf{A}_2 \label{form2}\\
\mathbf{Y}_3 =[[\mathbf{D}^{T_1}\mathbf{B}_3]^{T_2}\mathbf{A}_3]^{T_3}\mathbf{C}_3 \label{form3}\\
\mathbf{Y}_4 =[[\mathbf{D}^{T_1}\mathbf{B}_4]^{T_3}\mathbf{C}_4]^{T_2}\mathbf{A}_4 \label{form4}\\
\mathbf{Y}_5 =[[\mathbf{D}^{T_2}\mathbf{A}_5]^{T_1}\mathbf{B}_5]^{T_3}\mathbf{C}_5 \label{form5}\\
\mathbf{Y}_6 =[[\mathbf{D}^{T_2}\mathbf{A}_6]^{T_3}\mathbf{C}_6]^{T_1}\mathbf{B}_6 \label{form6}
\end{align}
\end{subequations}

We now provide the details of solutions to  the following two problems
\begin{itemize*}
  \item How to learn a base dictionary that is independent of subject and domains.
  \item Given an input face image and the base dictionary, how to obtain the sparse representation for the associated pose and illumination, and the domain-invariant sparse representation for the subject.
\end{itemize*}

\section{Compositional Dictionary Learning}
\label{sec:dadl}

In this section, we first show, given a domain base dictionary $\mathbf{D}$,  sparse coefficient matrices $\mathbf{\{A}_i\}_{i=1}^6$, $\mathbf{\{B}_i\}_{i=1}^6$ and $\mathbf{\{C}_i\}_{i=1}^6$ are { equal} across different equations in (\ref{eqn:6form-eq}). Then, we present  algorithms to learn a domain base dictionary $\mathbf{D}$, and perform domain-invariant sparse coding.

\subsection{Equivalence of Six Forms}

To learn a domain base dictionary $\mathbf{D}$, we first need to establish the following proposition.\\

\begin{proposition}
Given a domain base dictionary $\mathbf{D}$, matrices $\mathbf{\{A}_i\}_{i=1}^6$ in all six equations in (\ref{eqn:6form-eq}) are  { equal}, and so are matrices $\mathbf{\{B}_i\}_{i=1}^6$ and $\mathbf{\{C}_i\}_{i=1}^6$.
\end{proposition}

\begin{proof}
First we show matrices $\mathbf{B}_i$ in (\ref{form1}) and (\ref{form6}) are  { equal}.
$\mathbf{Y}_1$ and $\mathbf{Y}_6$ in Fig.~\ref{fig:6form} are different only in the row order.
 We assume a permutation matrix $\mathbf{P}_{16}$ will permutate the rows of $\mathbf{Y}_1$ into $\mathbf{Y}_6$, i.e., $\mathbf{P}_{16}\mathbf{Y}_1=\mathbf{Y}_6$.
Through a dictionary learning process, e.g., k-SVD \cite{Elad_KSVD}, we obtain a dictionary $\mathbf{D}_1$ and the associated sparse code matrix $\mathbf{B}_1$ for  $\mathbf{Y}_1$. $\mathbf{Y}_1$ can be reconstructed as $\mathbf{Y}_1 = \mathbf{D}_1\mathbf{B}_1$.
 We change the row order of $\mathbf{D}_1$ according to $\mathbf{P}_{16}$ without modifying the actual atom value as $\mathbf{D}_6 = \mathbf{P}_{16}\mathbf{D}_1$. We decompose $\mathbf{Y}_6$ using $\mathbf{D}_6$ as $\mathbf{Y}_6 = \mathbf{D}_6\mathbf{B}_6$, i.e., $\mathbf{P}_{16}\mathbf{Y}_1 = \mathbf{P}_{16}\mathbf{D}_1\mathbf{B}_6$, and we have $\mathbf{B}_1 = \mathbf{B}_6$.

Then we show that matrices $\mathbf{A}_i$, $\mathbf{B}_i$ and $\mathbf{C}_i$ in (\ref{form1}) and (\ref{form2}) are { equal}.
If we stack all the images from the same subject under the same pose but different illumination
as a single observation, we can consider $\mathbf{Y}_2=\mathbf{Y}_1^T$.
By assuming a bilinear model, we can represent $\mathbf{Y}_1$  as ${\mathbf{Y}_1=[\mathbf{D}_c\mathbf{A}_1]^T\mathbf{B}_1}$, and we have ${\mathbf{Y}_2=\mathbf{Y}_1^T=[\mathbf{D}_c^T\mathbf{B}_1]^T\mathbf{A}_1}$. As ${\mathbf{Y}_2=[\mathbf{D}_c^T\mathbf{B}_2]^T\mathbf{A}_2}$, $\mathbf{A}_i$ and $\mathbf{B}_i$ are { equal} in (\ref{form1}) and (\ref{form2}). As both equations share a bilinear map  $\mathbf{D}^{T_3}\mathbf{C}_i$,
with a common base dictionary $\mathbf{D}$, matrices $\mathbf{C}_i$ are also { equal} in (\ref{form1}) and (\ref{form2}).

Finally, we show matrices $\mathbf{A}_i$ and $\mathbf{C}_i$ in (\ref{form1}) and (\ref{form6}) are { equal}. We have shown in (\ref{form1}) and (\ref{form6}) that matrices $\mathbf{B}_i$ are { equal}. ${[[\mathbf{D}^{T_3}\mathbf{C}_1]^{T_2}\mathbf{A}_1]^{T_1}}$ and ${[[\mathbf{D}^{T_2}\mathbf{A}_6]^{T_3}\mathbf{C}_6]^{T_1}}$ are different only in the row order. We can use the bilinear model argument made above to easily show that matrices $\mathbf{A}_i$ and $\mathbf{C}_i$ are { equal} in (\ref{form1}) and (\ref{form6}).

Through the transitivity of equivalence, we can further show matrices $\mathbf{A}_i$ in all six equations in (\ref{eqn:6form-eq})  are equivalent, and so are matrices $\mathbf{B}_i$ and $\mathbf{C}_i$. We drop the subscripts in subsequent discussions and denote them as $\mathbf{A}$, $\mathbf{B}$ and $\mathbf{C}$.
\end{proof}

\subsection{Domain-invariant Sparse Coding}


\begin{algorithm}[t]
\KwIn{signals $\mathbf{Y}$, sparsity level $T_a$, $T_b$, $T_c$}
\KwOut{domain base dictionary $\mathbf{D}$}
\Begin{
\BlankLine
\textbf{Initialization stage:}\\
1. Initialize $\mathbf{B}$ by solving (\ref{form1}) via k-SVD\\
$\underset{\mathbf{D}_b,\mathbf{B}} \min\|\mathbf{Y}_1-\mathbf{D}_b\mathbf{B}\|_{F}^{2}, \; \mathrm{s.t.} \; \forall k \; \|\mathbf{b}_{k}\|_{o}\leq T_b,$ where $\mathbf{D}_b=[[\mathbf{D}^{T_3}\mathbf{C}]^{T_2}\mathbf{A}]^{T_1}$\\

\Repeat {convergence}{
2. apply $\mathbf{B}$ to (\ref{form1}) and solve via k-SVD
{ ($\mathbf{B}^{\dagger}=\mathbf{B}^{T}(\mathbf{B}^{T}\mathbf{B})^{-1}$)\\}
~~$\underset{\mathbf{D}_a,\mathbf{A}} \min\|{(\mathbf{Y}_1\mathbf{B}^{\dagger})}^{T_2}-\mathbf{D}_a\mathbf{A}\|_{F}^{2}, \; \mathrm{s.t.} \; \forall j \; \|\mathbf{a}_{j}\|_{o}\leq T_a,$ where ${\mathbf{D}_a=[\mathbf{D}}^{T_3}\mathbf{C}]^{T_2}$\\
3. apply $\mathbf{A}$ to (\ref{form4}) and solve via k-SVD\\
~~$\underset{\mathbf{D}_c,\mathbf{C}} \min\|{(\mathbf{Y}_4\mathbf{A}^{\dagger})}^{T_3}-\mathbf{D}_c\mathbf{C}\|_{F}^{2}, \; \mathrm{s.t.} \; \forall l \; \|\mathbf{c}_{l}\|_{o}\leq T_c,$ where ${\mathbf{D}_c=[\mathbf{D}}^{T_1}\mathbf{B}]^{T_3}$\\
4. apply $\mathbf{C}$ to (\ref{form5}) and solve via k-SVD\\
~~$\underset{\mathbf{D}_b,\mathbf{B}} \min\|{(\mathbf{Y}_5\mathbf{C}^{\dagger})}^{T_1}-\mathbf{D}_b\mathbf{B}\|_{F}^{2}, \; \mathrm{s.t.} \; \forall k \; \|\mathbf{b}_{k}\|_{o}\leq T_b,$ where ${\mathbf{D}_b=[\mathbf{D}}^{T_2}\mathbf{A}]^{T_1}$\\
}

5. Design the domain base dictionary: \\
$~~~~\mathbf{D} \leftarrow \mathbf{[D}^{T_2}\mathbf{A]A^{\dagger}}$\;

6. return $\mathbf{D}$\;
}
\caption{Domain base dictionary learning.}
\label{algo:basedict}
\end{algorithm}

\begin{algorithm}[ht]
\KwIn{an input face image $\mathbf{y}$, domain base dictionary $\mathbf{D}$, sparsity level $T_a$, $T_b$, $T_c$}
\KwOut{sparse representation vector for pose $\mathbf{a}$, illumination $\mathbf{c}$, subject $\mathbf{b}$}
\Begin{
\BlankLine
\textbf{Initialization stage:}\\
1. Initialize domain sparse code vector $\mathbf{a}$ and $\mathbf{c}$ with random values;

\textbf{Sparse coding stage:}\\
\Repeat {convergence}{
2. apply $\mathbf{a}$ and $\mathbf{c}$ to (\ref{form1}) and obtain $\mathbf{b}$ via { OMP},\\
~~~$\underset{\mathbf{b}} \min\|\mathbf{y}-\mathbf{[[D}^{T_3}\mathbf{c}]^{T_2}\mathbf{a}]^T\mathbf{b}\|_{2}^{2}, \; \mathrm{s.t.} \;  \|\mathbf{b}\|_{o}\leq T_b;$\\
3. apply $\mathbf{b}$ and $\mathbf{c}$ to (\ref{form4}) and obtain $\mathbf{a}$ via { OMP},\\
~~~$\underset{\mathbf{a}} \min\|\mathbf{y}-\mathbf{[[D}^{T_1}\mathbf{b}]^{T_3}\mathbf{c}]^T\mathbf{a}\|_{2}^{2}, \;\mathrm{ s.t.} \;  \|\mathbf{a}\|_{o}\leq T_a;$\\
4. apply $\mathbf{a}$ and $\mathbf{b}$ to (\ref{form5}) and obtain $\mathbf{c}$ via { OMP},\\
~~~$\underset{\mathbf{c}} \min\|\mathbf{y}-\mathbf{[[D}^{T_2}\mathbf{a}]^{T_1}\mathbf{b}]^T\mathbf{c}\|_{2}^{2}, \; \mathrm{s.t.} \;  \|\mathbf{c}\|_{o}\leq T_c;$\\

}

5. return \\
~~~domain-invariant subject sparse codes: $\mathbf{b}$,\\
~~~pose sparse codes: $\mathbf{a}$,\\
~~~illumination sparse codes: $\mathbf{c}$;

}
\caption{Domain-invariant sparse coding.}
\label{algo:coding}
\end{algorithm}

As matrices $\mathbf{A}$, $\mathbf{B}$ and $\mathbf{C}$ are { equal} across all six forms in (\ref{eqn:6form-eq}), we propose to learn the base dictionary $\mathbf{D}$ using Algorithm \ref{algo:basedict} given below.
{
The domain dictionary learning in Algorithm~\ref{algo:basedict} optimizes the following objective function,
\begin{align} \label{dlobj}
 \underset{\mathbf{D}, \mathbf{A}, \mathbf{B}, \mathbf{C}}  \min & \|\mathbf{Y}_1-[[\mathbf{D}^{T_3}\mathbf{C}]^{T_2}\mathbf{A}]^{T_1}\mathbf{B}\|_{F}^{2}, \; \\ \nonumber
& \mathrm{s.t.}  \;
 \forall j \; \|\mathbf{a}_{j}\|_{o}\leq T_a,
 \forall k \; \|\mathbf{b}_{k}\|_{o}\leq T_b,
 \forall l \; \|\mathbf{c}_{l}\|_{o}\leq T_c,
\end{align}
where  $T_a$, $T_b$, and $T_c$ specify the sparsity level, i.e., the maximal number of non-zero values in a sparse vector.
For simplicity and efficiency,  we optimize (\ref{dlobj}) as  a sequence of  dictionary learning subproblems.
More specifically,  we first let $\mathbf{D}_b=[[\mathbf{D}^{T_3}\mathbf{C}]^{T_2}\mathbf{A}]^{T_1}$, and perform regular sparse dictionary learning to solve
\[
\underset{\mathbf{D}_b,\mathbf{B}} \min\|\mathbf{Y}_1-\mathbf{D}_b\mathbf{B}\|_{F}^{2}, \; \mathrm{s.t.} \; \forall k \; \|\mathbf{b}_{k}\|_{o}\leq T_b.
\]
We then use the obtained $\mathbf{B}$  to seek an update to $\mathbf{D}_b$ to minimize the same error $\|\mathbf{Y}_1-\mathbf{D}_b\mathbf{B}\|_{F}^{2}$.
Taking the derivative with respect to $\mathbf{D}_b$, we obtain $(\mathbf{Y}_1-\mathbf{D}_b\mathbf{B})\mathbf{B}^{T}=0$, leading to the updated $\mathbf{D}_b$ as
$ \mathbf{Y}_1 \mathbf{B}^{\dagger}=\mathbf{Y}_1 \mathbf{B}^{T}(\mathbf{B}^{T}\mathbf{B})^{-1}.$
As $\mathbf{D}_b=[[\mathbf{D}^{T_3}\mathbf{C}]^{T_2}\mathbf{A}]^{T_1}$, we now use updated $\mathbf{D}_b$ to obtain $\mathbf{D}_a$ and $\mathbf{A}$ as
\[
\underset{\mathbf{D}_a,\mathbf{A}} \min\|{(\mathbf{Y}_1\mathbf{B}^{\dagger})}^{T_2}-\mathbf{D}_a\mathbf{A}\|_{F}^{2}, \; \mathrm{s.t.} \; \forall j \; \|\mathbf{a}_{j}\|_{o}\leq T_a,
\]
 where ${\mathbf{D}_a=[\mathbf{D}}^{T_3}\mathbf{C}]^{T_2}$. Then we fix $\mathbf{A}$ to update $\mathbf{D}_a$, and solve for
$\mathbf{D}_c$ and $\mathbf{C}$, and so on.
}

Algorithm 1 is designed as an iterative method, and each iteration consists of several typical sparse dictionary learning problems. Thus, this algorithm is flexible and can rely on any sparse dictionary learning methods. We adopt the highly efficient dictionary learning method, k-SVD \cite{Elad_KSVD}.
It is noted that we can easily omit one domain aspect through dictionary ``marginalization". For example, after learning the { base}  dictionary $\mathbf{D}$, we can marginalize over illumination sparse codes matrix $\mathbf{C}$ and  adopt ${[\mathbf{D}^{T_3}\mathbf{C}]^{T_2}}$ as the base dictionary for pose domains only.

With the learned base dictionary $\mathbf{D}$, we can perform domain-invariant sparse coding as shown in Algorithm~\ref{algo:coding},
{
which minimizes the following objective function for a fixed $\mathbf{D}$,
\begin{align} \label{ompobj}
\underset{ \mathbf{a}, \mathbf{b}, \mathbf{c}} \min &\|\mathbf{y}-[[\mathbf{D}^{T_3}\mathbf{c}]^{T_2}\mathbf{a}]^{T_1}\mathbf{b}\|_{F}^{2}, \\ \nonumber
&\; \mathrm{s.t.}  \;
\; \|\mathbf{a}\|_{o}\leq T_a,
 \; \|\mathbf{b}\|_{o}\leq T_b,
\; \|\mathbf{c}\|_{o}\leq T_c.
\end{align}
The $l_0$ norm minimization involved here  is  NP-hard and usually solved using  greedy pursuit algorithms, such as  basis pursuit,  orthogonal matching pursuit (OMP) \cite{omp,greedgood}, to
represent a signal with the best linear combination of $t$ atoms from a dictionary, where $t$ is the sparsity.
We adopt OMP in the paper. OMP is a greedy algorithm that iteratively selects  the dictionary atom best correlated with the residual; and then it produces a new approximant by projecting the signal onto atoms  already been selected.

With a  base dictionary $\mathbf{D}$ learned using Algorithm~\ref{algo:basedict},  a face image can be decomposed using Algorithm~\ref{algo:coding} into sparse representations $\mathbf{a}$ for the associated pose and  $\mathbf{c}$ for illumination, and a domain-invariant sparse representation $\mathbf{b}$ for the subject.
While minimizing  (\ref{dlobj}) in Algorithm~\ref{algo:basedict}, we obtain the learned domain dictionary $\mathbf{D}$, and model codes $\mathbf{A}$, $\mathbf{B}$, and $\mathbf{C}$.  Each column of $\mathbf{A}$ denotes the sparse representation assigned to a particular pose in the training.
When a training pose shown at testing,  the decomposed pose code $\mathbf{a}$ using Algorithm~\ref{algo:coding} converges to the respective column in $\mathbf{A}$.
As shown later,  testing poses unseen at training are converged to a sparse linear combination of known poses in a consistent way.  We can observe similar convergence for both subject codes and illumination codes.

Convergence of Algorithms~\ref{algo:basedict} and~\ref{algo:coding} can be established using the convergence results of k-SVD discussed in \cite{Elad_KSVD}.
Although  both algorithms optimize a single objective function, the convergence  depends on the success of greedy pursuit algorithms involved in each iteration step. We have observed empirical convergence for both Algorithm~\ref{algo:basedict} and~\ref{algo:coding} in all the experiments reported below.

During training, the domain dictionary learning consists of multiple k-SVD procedures, and the complexity of k-SVD  is analyzed in details in \cite{ksvd2}. The complexity of Algorithm 2 is more critical,  as domain-invariant sparse coding is usually performed at  testing. As shown later, it  usually takes about 10 iterations for Algorithm 2 to converge, and each iteration consists of three OMP operations. As discussed in \cite{ksvd2, omp-comp},  different implementations of OMP have different complexities.  Considering a signal of dimension $m$ with assumed sparsity $t$, and a dictionary of $N$ atoms,  OMP implemented using the QR Decomposition has the complexity of $Nt + mt + t^2$.
}

\section{Experimental Evaluation}
\label{sec:experiment}

\begin{figure} [ht]
\centering
\subfloat[Illumination variation] {\label{fig:pie_illum} \includegraphics[angle=0, height=0.06\textwidth, width=.5\textwidth]{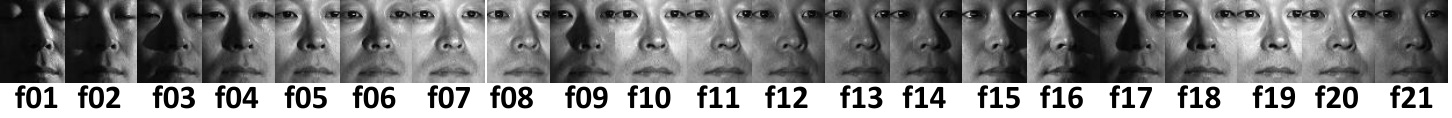}}
\\ \subfloat[Pose variation] {\label{fig:pie_pose} \includegraphics[angle=0, height=0.18\textwidth, width=.5\textwidth]{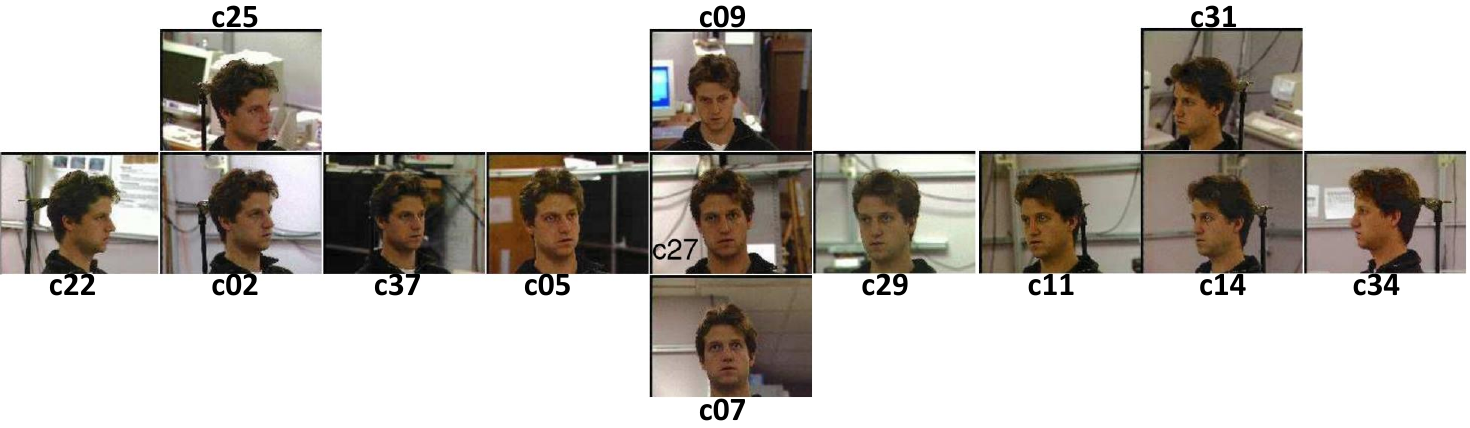} \hspace{5pt}}
\caption{Pose and illumination variation in the PIE dataset.}
\label{fig:pie}
\end{figure}

This section presents experimental evaluations on two public face datasets: the CMU PIE dataset \cite{pie} and the Extended YaleB dataset \cite{yaleb}. The PIE dataset consists of 68 subjects imaged simultaneously under 13 different poses and 21 lighting conditions, as shown in Fig.~\ref{fig:pie}.
The Extended YaleB dataset contains 38 subjects with near frontal pose under 64 lighting conditions.
$64 \times 48$ sized images are used in the domain composition experiments in Section~\ref{compexp} for clearer visualization. In the remaining experiments, all the face images are resized to $32 \times 24$.  The proposed Compositional Dictionary learning method is refereed to as CDL in subsequent discussions.

Experimental evaluation is summarized as follows:
Section~\ref{sec:dict} provides the learning configurations for all base dictionaries used.
 The convergence of domain-invariant sparse coding is illustrated in Section~\ref{convergence}.
Section~\ref{compexp} demonstrates how domain composition is used for pose alignment and illumination normalization.
Domina-invariant subject representation is adopted  for cross-domain recognition in Section~\ref{sec:facerec}.
Section~\ref{errorcode} shows that the proposed method is more robust for multilinear decompostion than the tensor-based method. Domain estimation is demonstrated in  Section~\ref{domainEst}.

\subsection{Learned Domain Base Dictionaries}
\label{sec:dict}

In our experiments, four different domain base dictionaries $\mathbf{D}_{10}$, $\mathbf{D}_{4}$, $\mathbf{D}_{34}$, and $\mathbf{D}_{32}$ are learned. We explain here the configurations for each base dictionary.

\begin{itemize*}
\item $\mathbf{D}_{4}$: This dictionary is learned from the PIE dataset by using 68 subjects in 4 poses under 21 illumination conditions. The four training poses to the dictionary are  $c02$, $c07$, $c09$ and $c14$ poses shown in Fig.~\ref{fig:pie}.
    The dimensions of coefficient vectors for subject, pose and illumination are 68, 4 and 9. The respective coefficient sparsity values, i.e., the maximal number of non-zero coefficients, are 20, 4 and 9.
    { Note that there is no defined way to specify the sparsity value, and we manually specify sparsity  to make each coefficient vector around $\frac{1}{3}$ to $\frac{1}{2}$ full.}
\item $\mathbf{D}_{10}$: This dictionary is learned from the PIE dataset by using 68 subjects in 10 poses under all 21 illumination conditions. The three unknown poses to the dictionary are $c27$ (frontal), $c05$ (side) and $c22$ (profile) poses.
    The dimensions of coefficient vectors for subject, pose and illumination are 68, 10 and 9. The respective coefficient sparsity values are 20, 8 and 9.
\item $\mathbf{D}_{34}$: This dictionary is learned from the PIE dataset by using the first 34 subjects in 13 poses under 21 illumination conditions.
    The dimensions of coefficient vectors  for subject, pose and illumination are 34, 13 and 9. The respective coefficient sparsity values are 12, 8 and 9.
\item $\mathbf{D}_{32}$: This dictionary is learned from the Extended YaleB dataset by using 38 subjects under 32 randomly selected lighting conditions.
    The dimensions of coefficient vectors for subject and illumination are 38, and 32. The respective coefficient sparsity values are 20 and 20.
\end{itemize*}

The choice of the above 4 dictionary configurations is explained as follows: usually, a common challenging setup for the PIE dataset is to classify subjects in three poses: \emph{frontal} ($c27$), \emph{side} ($c05$) and \emph{profile} ($c22$). Given 13 poses in PIE, we keep the remaining 10 poses to learn $\mathbf{D}_{10}$;
We further experiment with fewer samples, e.g., a subset of the remaining 10 poses, to learn $\mathbf{D}_4$; Given 68 subjects in PIE, we learn $\mathbf{D}_{34}$ using half of the subjects; Given 64 illumination conditions in the Extended YaleB data, we learn $\mathbf{D}_{32}$ using half of the lighting conditions.

\subsection{Convergence of Domain-invariant Sparse Coding}
\label{convergence}

\begin{figure*} [ht]
\centering
 \subfloat[Sparse codes decomposed over $\mathbf{D}_{10}$ after $1$ iteration.] {\label{fig:coding1_1} \includegraphics[angle=0, height=0.15\textwidth, width=.8\textwidth]{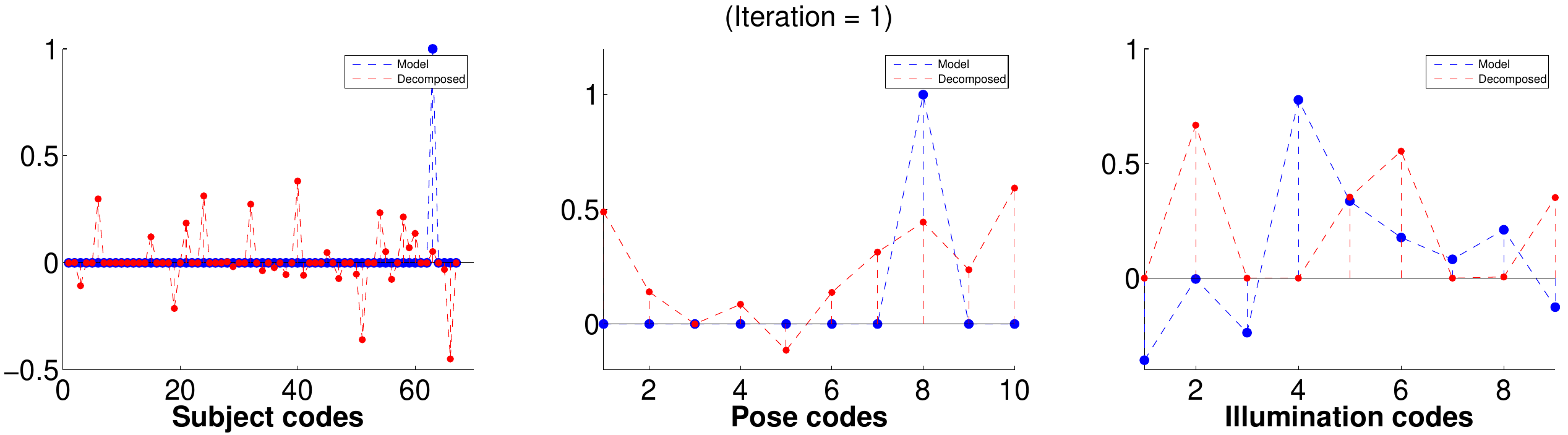} \hspace{10pt}} \\
\subfloat[Sparse codes decomposed over $\mathbf{D}_{10}$ after $2$ iterations.] {\label{fig:coding1_2} \includegraphics[angle=0, height=0.15\textwidth, width=.8\textwidth]{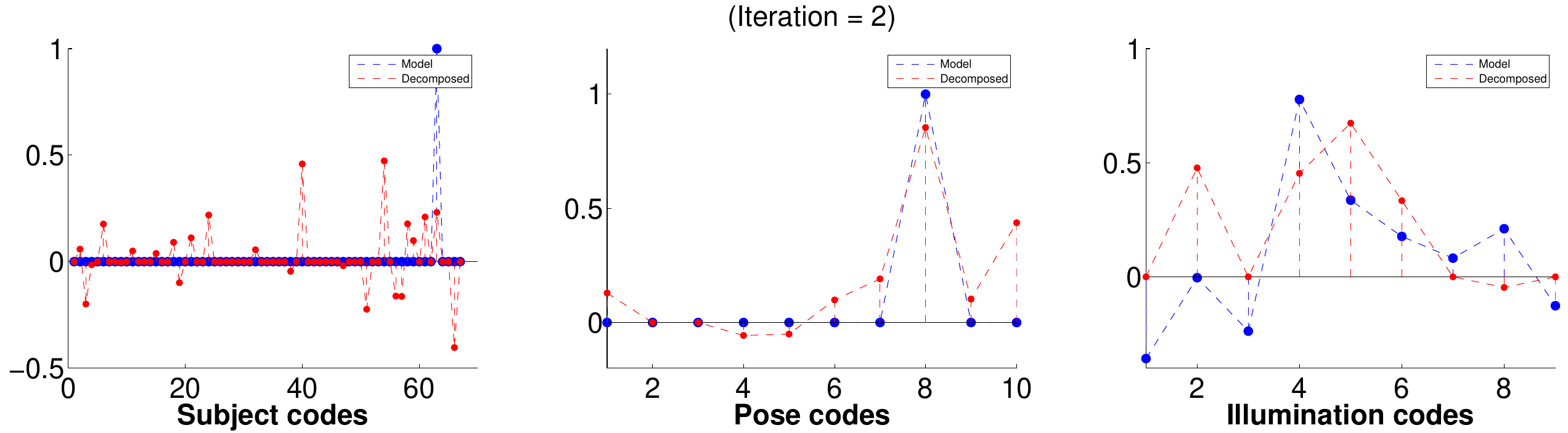}}\\
 \subfloat[Sparse codes decomposed over $\mathbf{D}_{10}$ after $100$ iterations.] {\label{fig:coding1_100} \includegraphics[angle=0, height=0.15\textwidth, width=.8\textwidth]{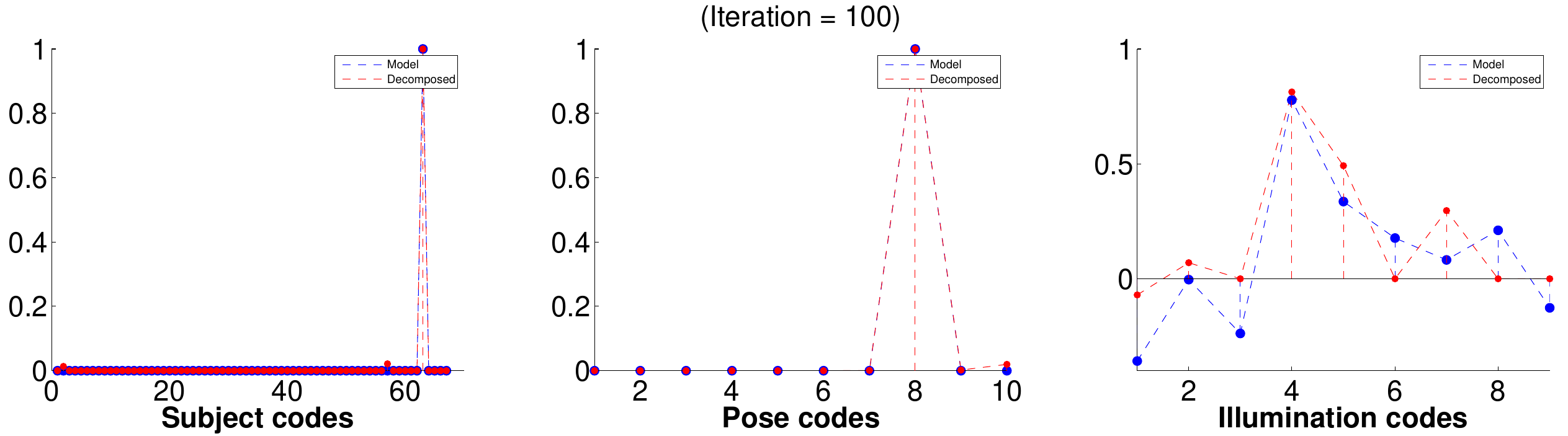}\hspace{10pt}}\\
\subfloat[Sparse codes decomposed over a \emph{Tensor k-SVD} dictionary \cite{tensor-ksvd} after $100$ iterations.] {\label{fig:ncoding1_100} \includegraphics[angle=0, height=0.15\textwidth, width=.8\textwidth]{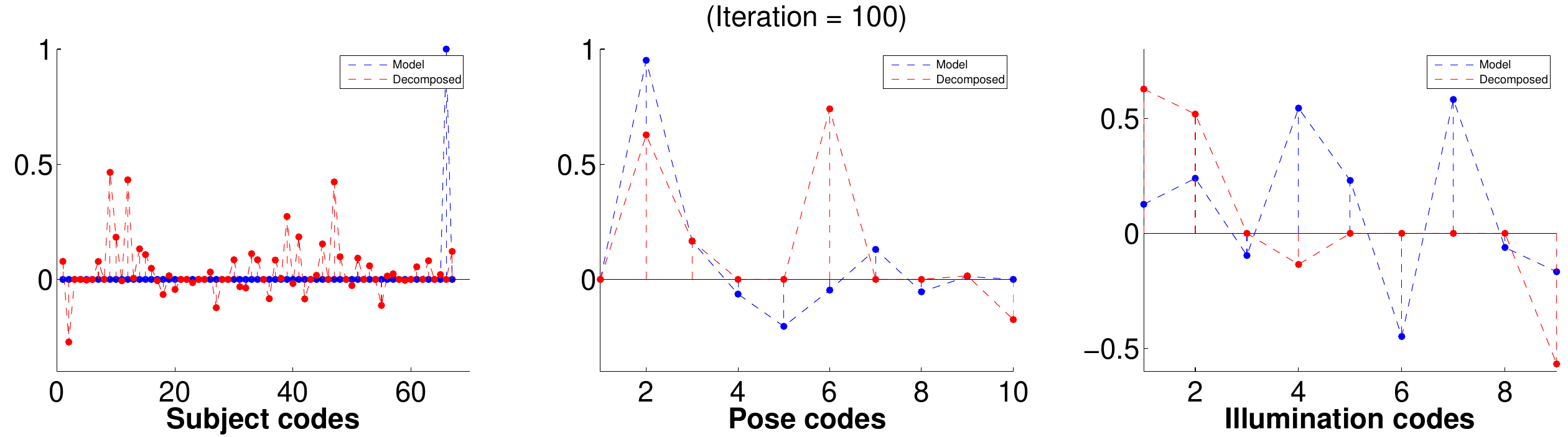}}
\caption{Domain-invariant sparse coding (Algorithm~\ref{algo:coding}) of the face image $(s43, c29, f05)$, i.e., subject $s43$ in pose $c29$ and illumination $f05$, over a domain base dictionary. $c29$ is a known pose to $\mathbf{D}_{10}$.
{
In (a)-(c), the decomposed sparse codes (color red) converge to the model codes (color blue) when the base dictionary is learned using Algorithm~\ref{algo:basedict}; and such convergence is not warranted in (d), when the base dictionary is not from Algorithm~\ref{algo:basedict}.
}
}
\label{fig:coding1}
\end{figure*}

\begin{figure*} [ht]
\centering
 \subfloat[Sparse codes of $(s43, c27, f13)$ decomposed over $\mathbf{D}_{10}$ after $100$ iterations.] {\label{fig:coding2_1} \includegraphics[angle=0, height=0.15\textwidth, width=.8\textwidth]{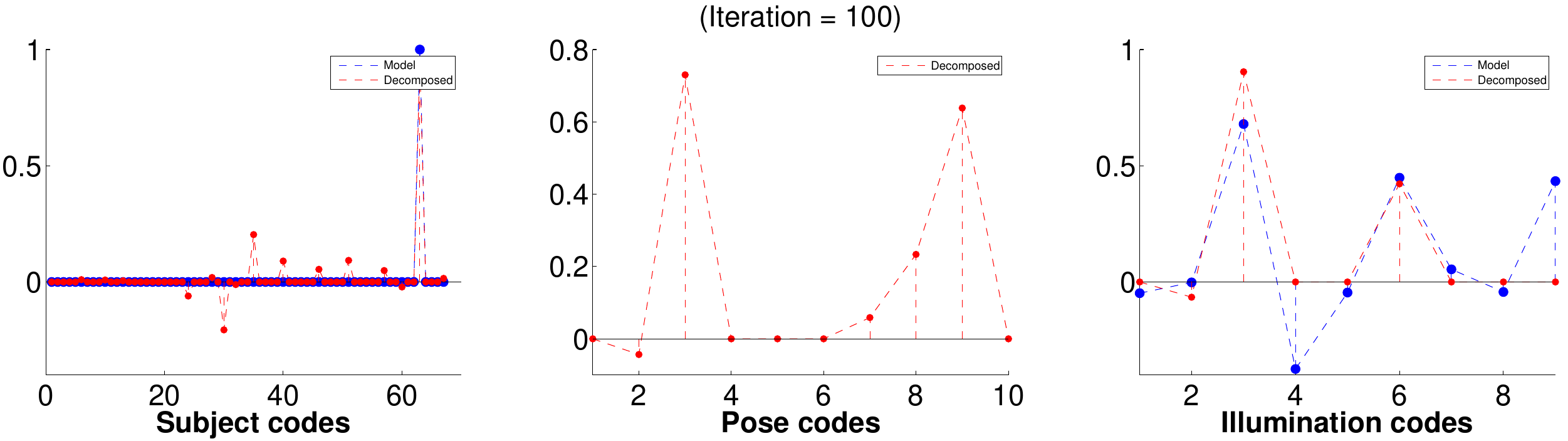}}\\
 \subfloat[Sparse codes of $(s01, c27, f13)$ decomposed over $\mathbf{D}_{10}$ after $100$ iterations.] {\label{fig:coding2_2} \includegraphics[angle=0, height=0.15\textwidth, width=.8\textwidth]{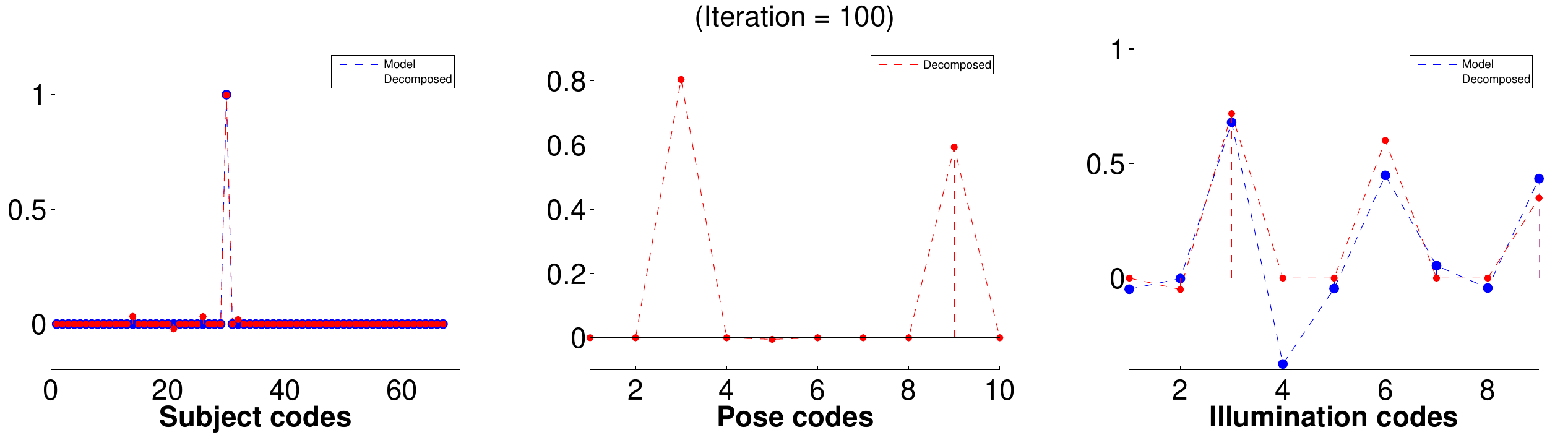}}
\caption{Domain-invariant sparse coding (Algorithm~\ref{algo:coding}) of face images $(s43, c27, f13)$ and $(s01, c27, f13)$ over the domain base dictionary $\mathbf{D}_{10}$. $c27$ is an unknown pose to $\mathbf{D}_{10}$.
The decomposed subject and illumination codes (color red) converge to the learned model codes (color blue). Note that the unknown pose $c27$ is represented as a sparse linear combination of known poses in a consistent way.}
\label{fig:coding2}
\end{figure*}

\begin{figure*} [ht]
\centering
 \subfloat[Sparse decomposition of $(s43, c29, f05)$.] {\label{fig:converge_1} \includegraphics[angle=0, height=0.18\textwidth, width=.3\textwidth]{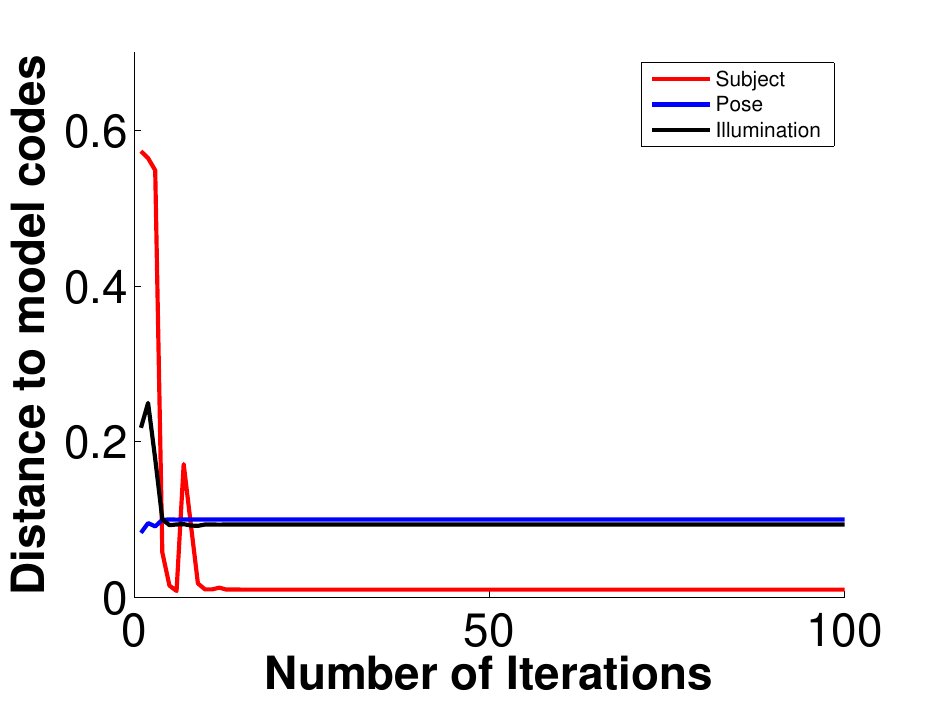}}
  \subfloat[Sparse decomposition of $(s43, c27, f13)$.] {\label{fig:converge_2} \includegraphics[angle=0, height=0.18\textwidth, width=.3\textwidth]{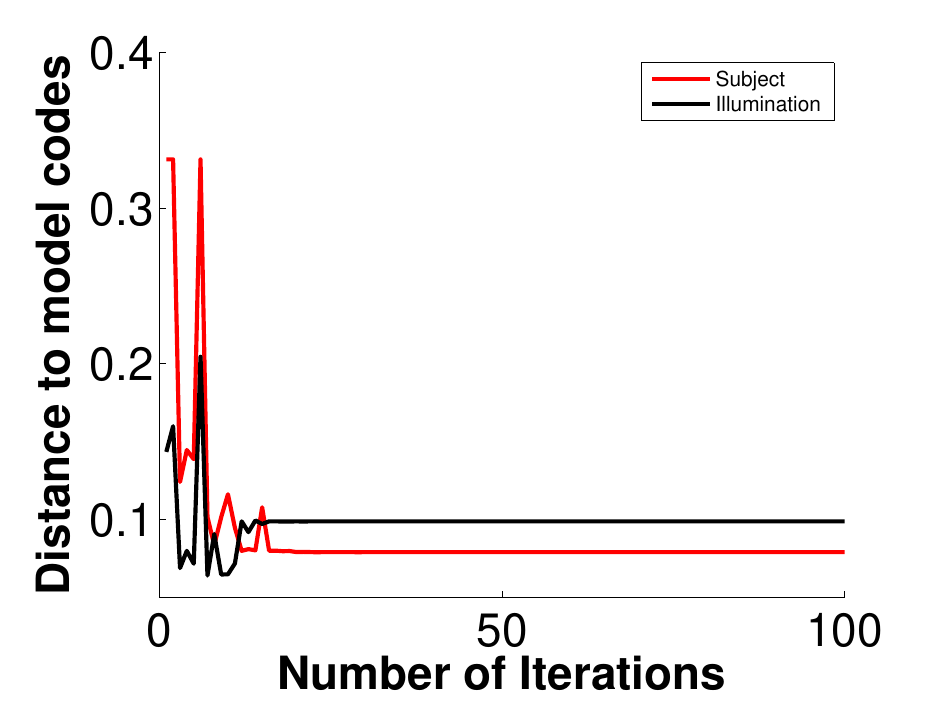}}
    \subfloat[Sparse decomposition of $(s01, c27, f13)$.] {\label{fig:converge_3} \includegraphics[angle=0, height=0.18\textwidth, width=.3\textwidth]{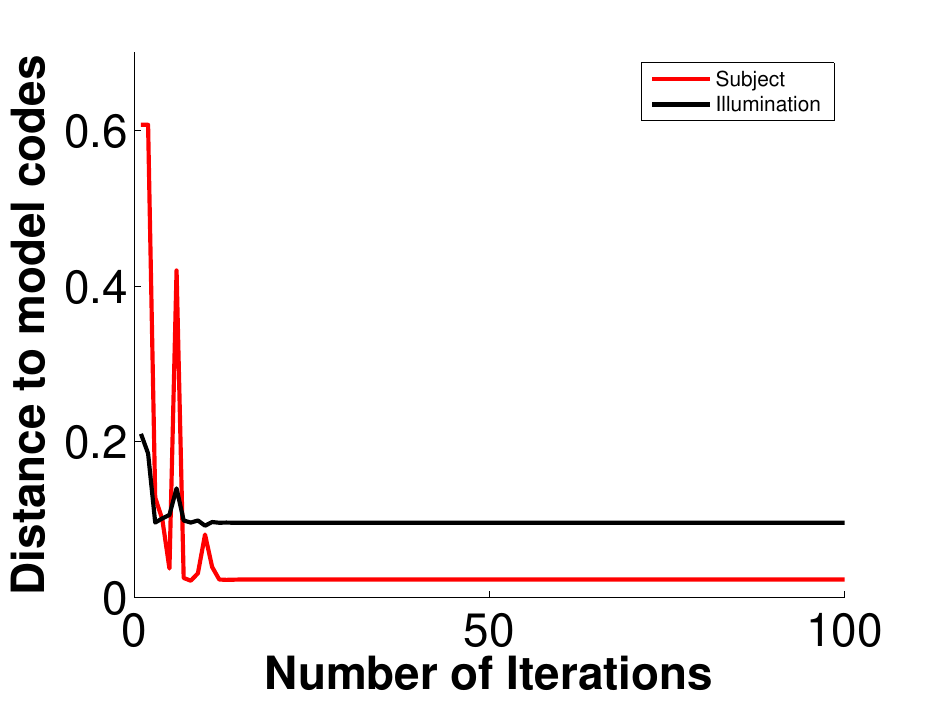}}
\caption{Convergence of domain-invariant sparse coding in Algorithm~\ref{algo:coding}. }
\label{fig:converge}
\end{figure*}

We demonstrate here the convergence of the proposed domain-invariant sparse coding in Algorithm~\ref{algo:coding} over a base dictionary learned using Algorithm~\ref{algo:basedict}.
We first learn the domain base dictionary $\mathbf{D}_{10}$ using Algorithm~\ref{algo:basedict}, and also obtain the associated domain matrices { (learned model codes)} $\mathbf{A}$, $\mathbf{B}$ and $\mathbf{C}$.
The matrix $\mathbf{A}$ consists of $10$ columns and each column is a unique sparse representation for one of the 10 poses. The matrix $\mathbf{B}$ consists of $68$ columns, and each column describes one of the $68$ subjects.  The matrix $\mathbf{C}$ consists of $21$ columns, and each column describes one of the $21$ illumination conditions. We observe no significant reconstruction improvements from the learned base dictionary after $2$ iterations of Algorithm~\ref{algo:basedict}.

Given a face image $(s43, c29, f05)$, i.e., subject $s43$ in pose $c29$ under illumination $f05$,
Fig.~\ref{fig:coding1_1}, \ref{fig:coding1_2} and \ref{fig:coding1_100} show the decomposed sparse representations for subject $s43$, pose $c29$ and illumination $f05$ after $1$, $2$ and $100$ iterations of Algorithm~\ref{algo:coding} respectively. We can notice that the decomposed sparse codes (color red) converge to
the learned model codes (color blue) in $\mathbf{A}$, $\mathbf{B}$ and $\mathbf{C}$. As shown in Fig.~\ref{fig:converge_1}, we observe convergence after 4 iterations.

\cite{tensor-ksvd} proposed a \emph{Tensor k-SVD} method, which is similar to Tensorfaces but replaces the $N$-mode SVD with k-SVD to perform multilinear sparse decomposition. Using the Tensor k-SVD method, we are able to learn a Tensor k-SVD dictionary and the associated domain matrices.
As the Tensor k-SVD method is designed for data compression, it is not discussed in \cite{tensor-ksvd} how to decompose a single image into separate { sparse}   coefficient vectors over such learned Tensor k-SVD dictionary.
We adopt a learned Tensor k-SVD dictionary as the base dictionary for domain-invariant sparse coding using Algorithm~\ref{algo:coding}.
As shown in Fig.\ref{fig:ncoding1_100}, the decomposed sparse codes do not converge well to the learned model codes. It indicates that Algorithm~\ref{algo:coding} performs an inconsistent decomposition over the Tensor k-SVD dictionary.
Therefore, a base dictionary learned from Algorithm~\ref{algo:basedict} is required by the proposed domain-invariant sparse coding in Algorithm~\ref{algo:coding} to enforce a consistent multilinear sparse decomposition.

We further decompose face images $(s43, c27, f13)$ and $(s01, c27, f13)$ over $\mathbf{D}_{10}$.
As shown in Fig.~\ref{fig:coding2}, even when pose $c27$ is unknown to $\mathbf{D}_{10}$,  the decomposed sparse codes for subjects $s43$ and $s01$, and illumination $f13$ still converge to the learned models.
By comparing the pose codes in Fig.~\ref{fig:coding2_1} and \ref{fig:coding2_2}, we notice that the unknown pose $c27$ is represented as a sparse linear combination of known poses in a consistent way.
 Given the non-optimality of the greedy OMP adopted in each iteration \cite{greedgood},  we still observe convergence after about 10 iterations for both cases, as shown in Fig.\ref{fig:converge_2} and Fig.\ref{fig:converge_3}.

\subsection{Domain Composition}
\label{compexp}

\begin{figure*} [ht]
\centering
 \subfloat[Composition using base dictionary $\mathbf{D}_{34}$. s01 is a known subject to $\mathbf{D}_{34}$. { c27 and f05 are extracted from an unknown subject s43}.] {\label{fig:d34known} \includegraphics[angle=0, height=0.18\textwidth, width=1\textwidth]{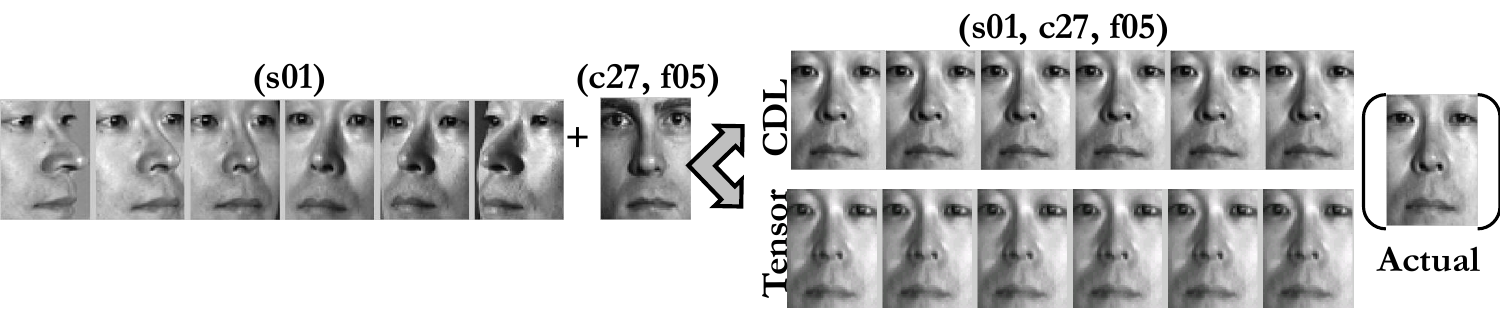}}
\\ \subfloat[Composition using base dictionary $\mathbf{D}_{34}$. s43 is an unknown subject to $\mathbf{D}_{34}$.] {\label{fig:d34unknown} \includegraphics[angle=0, height=0.27\textwidth, width=1\textwidth]{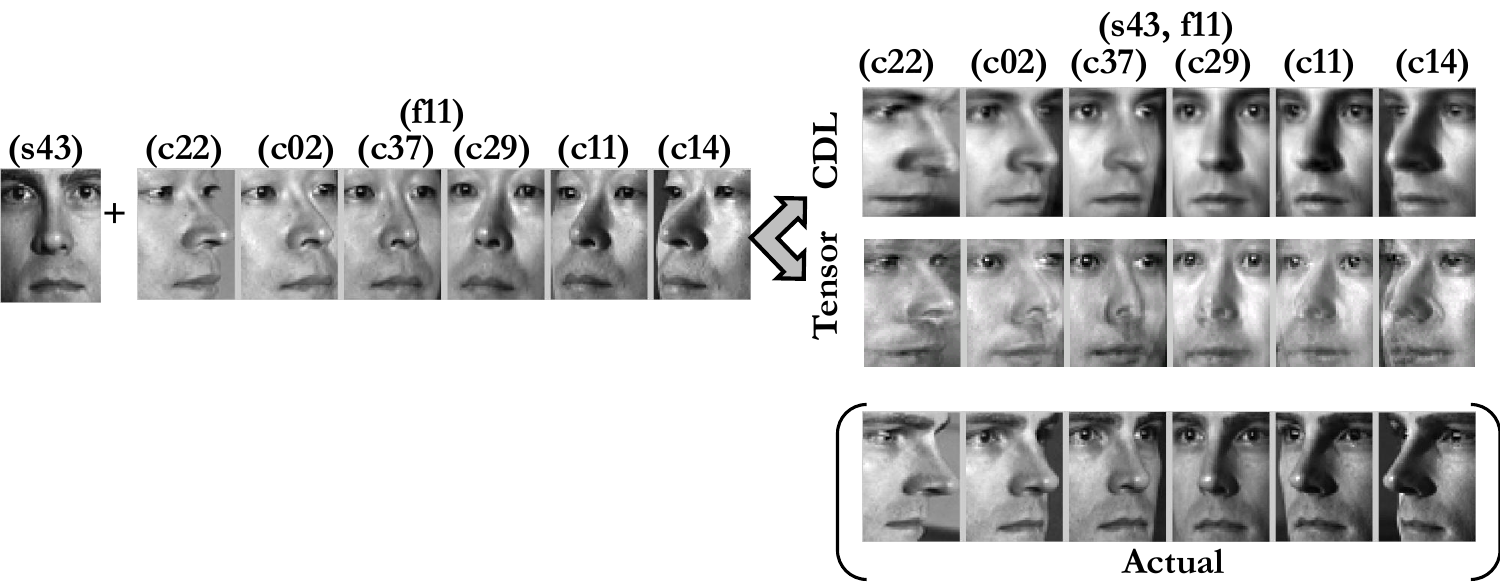}}
\\ \subfloat[Composition using base dictionary $\mathbf{D}_{10}$. c22, c05 and c27 are unknown poses to $\mathbf{D}_{10}$.] {\label{fig:d68unknown} \includegraphics[angle=0, height=0.18\textwidth, width=.8\textwidth]{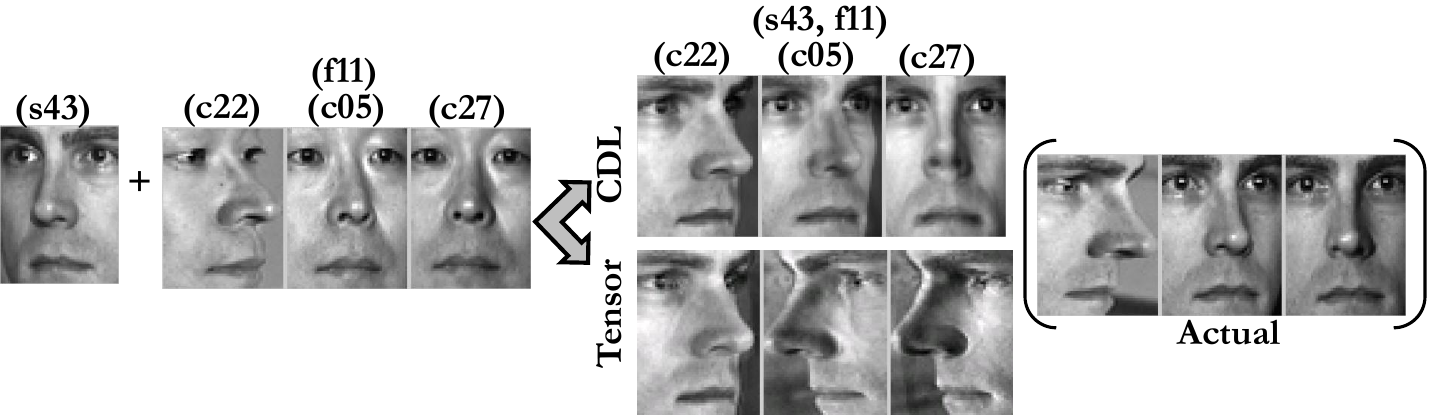}}
\caption{Pose alignment through domain composition. In each corresponding Tensorfaces experiment, we adopt the same training data and sparsity values used for the CDL base dictionary for a fair comparison. When a subject or a pose is unknown to the training data, the proposed CDL method provides significantly more accurate reconstruction to the ground truth images. }
\label{fig:pose_align}
\end{figure*}

\begin{figure*} [ht]
\centering
 \subfloat[Composition using base dictionary $\mathbf{D}_{34}$. s28 is a known subject to $\mathbf{D}_{34}$.] {\label{fig:d34knownlight} \includegraphics[angle=0, height=0.18\textwidth, width=.91\textwidth]{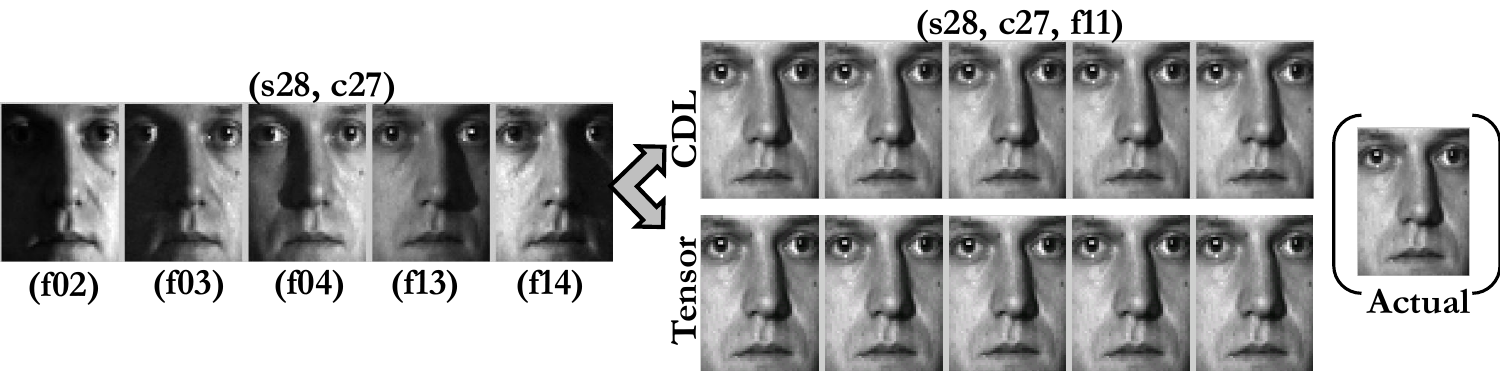} }
\\ \subfloat[Composition using base dictionary $\mathbf{D}_{34}$. s43 is an unknown subject to $\mathbf{D}_{34}$.] {\label{fig:d34unknownlight} \includegraphics[angle=0, height=0.18\textwidth, width=.91\textwidth]{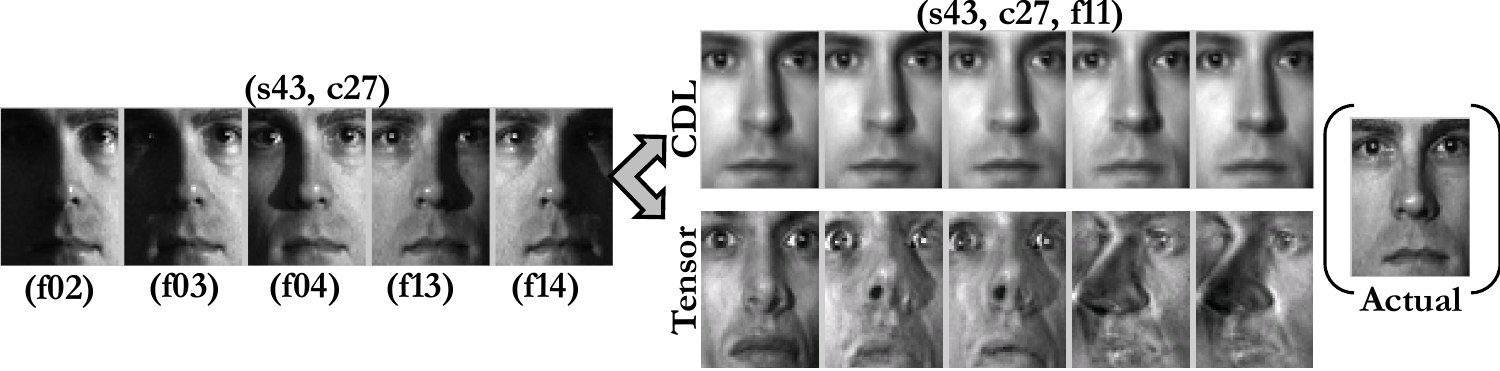}}
\caption{Illumination normalization through domain composition. In each corresponding Tensorfaces experiment, we adopt the same training data and sparsity values used for the CDL base dictionary for a fair comparison. When a subject is unknown to the training data, the proposed CDL method provides significantly more accurate reconstruction to the ground truth images.}
\label{fig:light_remove}
\end{figure*}

Using the proposed trilinear sparse decomposition over a base dictionary as illustrated in Algorithm~\ref{algo:coding}, we extract from a face image the respective sparse representations for subject, pose and illumination. We can then translate a subject to a different pose and illumination by composing the corresponding subject and domain sparse codes over the base dictionary.
As discussed in Sec.~\ref{sec:tensor}, Tensorfaces also enable the decomposition of a face image into separate coefficients for the subject, pose and illumination through exhaustive projections and matchings. We adopt the Tensorfaces method here for a fair comparison in our domain composition experiments.

\subsubsection{Pose Alignment}
In Fig.~\ref{fig:d34known}, the base dictionary $\mathbf{D}_{34}$ is used in the CDL experiments. To enable a fair comparison, we adopt the same training data and sparsity values for $\mathbf{D}_{34}$ in the corresponding Tensorfaces experiments.
Given faces from subject s01 under different poses, where both the subject and poses are present in the training data, we extract the subject (sparse) codes for $s01$ from each of them. Then we extract the pose codes for $c27$ (frontal) and the illumination codes for $f05$ from an image of subject $s43$. It is noted that, for such \emph{known subject} cases, the composition $(s01, c27, f05)$ through both CDL and Tensorfaces provides  good reconstructions to the ground truth image. The reconstruction using CDL is clearer than the one using Tensorfaces.

In Fig.~\ref{fig:d34unknown}, we first extract the subject codes for $s43$, which is an unknown subject to $\mathbf{D}_{34}$. Then we extract the pose codes and the illumination codes from the set of images of $s01$ in Fig.~\ref{fig:d34known}.
In this \emph{unknown subject} case, the composition using our CDL method provides significantly more accurate reconstruction to the groundtruth images than the Tensorfaces method.
The central assumption in the literature on sparse representation for faces is that the test face image should be represented in terms of training images of the same subject \cite{Wright09}, \cite{close-loop}. As $s43$ is unknown to $\mathbf{D}_{34}$, therefore, it is expected that the reconstruction of the subject information is through a linear combination of other known subjects, which is an approximation but not exact.

In Fig.~\ref{fig:d68unknown}, the base dictionary $\mathbf{D}_{10}$ is used in the CDL experiments, and the same training data and sparsity values for $\mathbf{D}_{10}$ are used in the corresponding Tensorfaces experiments. We first extract the subject codes for $s43$.  Then we extract the pose codes for pose $c22$, $c05$ and $c27$, which are unknown poses to the training data. Through domain composition, for such \emph{unknown pose} cases, we obtain more acceptable reconstruction to the actual images using CDL than Tensorfaces.
This indicates that, using the proposed CDL method, an unknown pose can be much better approximated in terms of a set of observed poses.

\subsubsection{Illumination Normalization}
In Fig.~\ref{fig:d34knownlight}, we use frontal faces from subject $s28$, which is known to  $\mathbf{D}_{34}$,  under different illumination conditions. For each image,  we first isolate the codes for subject, pose and illumination, and then replace the illumination codes with the one for $f11$. If $f11$ is observed in the training data, the illumination codes for $f11$ can be obtained during training. Otherwise, the illumination codes for f11 can be extracted from { a face image of any subjects} under $f11$ illumination. It is shown in Fig.~\ref{fig:d34knownlight} that, for such \emph{known subject} cases, after removing the illumination variation, we can obtain a reconstructed image close to the ground truth image using both CDL and Tensorfaces.

Subject $s43$ in Fig.~\ref{fig:d34unknownlight} is unknown to $\mathbf{D}_{34}$. The composed images from CDL exhibit significantly more accurate subject, pose and illumination reconstruction than Tensorfaces. As discussed before, the reconstruction to the subject here is only an approximation but not exact.

\subsection{Pose and Illumination Invariant Face Recognition}
\label{sec:facerec}

\subsubsection{Classifying PIE 68 Faces using $\mathbf{D}_{4}$ and $\mathbf{D}_{10}$}

\begin{figure*} [ht]
\centering
 \subfloat[Gallery: profile. Probe: frontal.] {\label{fig:acc-PF} \includegraphics[angle=0, height=0.18\textwidth, width=.33\textwidth]{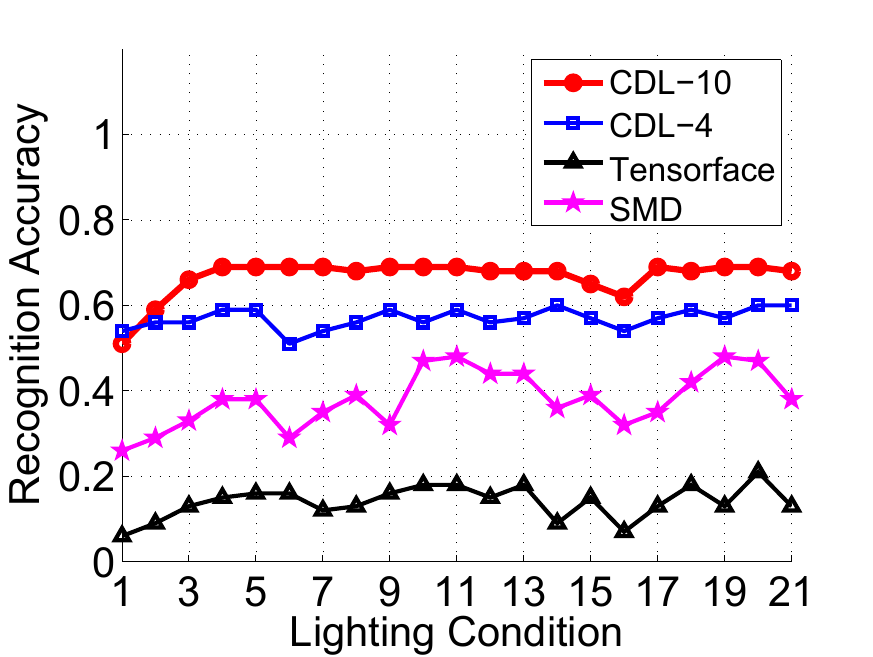} }
  \subfloat[Gallery: profile. Probe: side.] {\label{fig:acc-PS} \includegraphics[angle=0, height=0.18\textwidth, width=.33\textwidth]{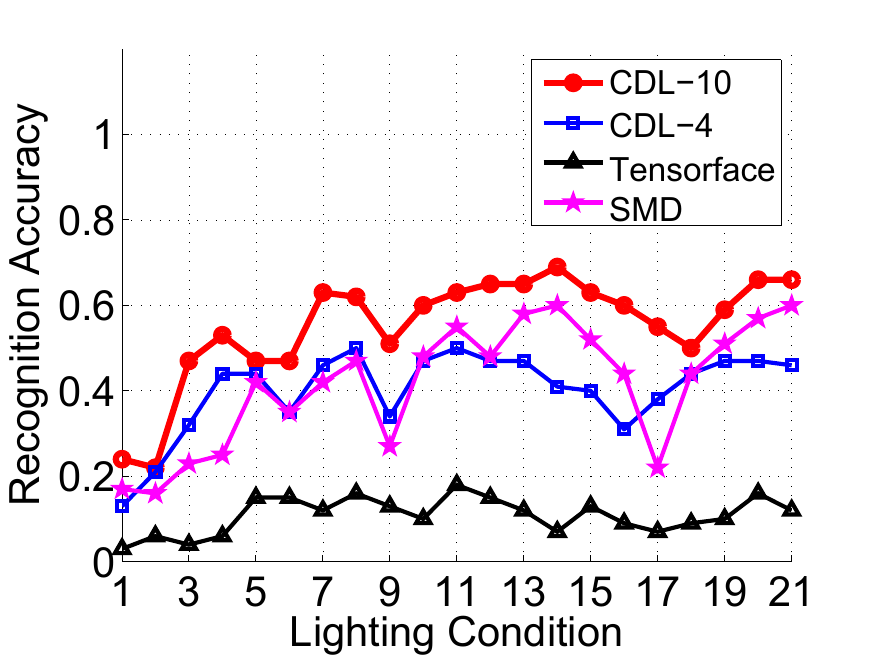} }
   \subfloat[Gallery: frontal. Probe: side.] {\label{fig:acc-FS} \includegraphics[angle=0, height=0.18\textwidth, width=.33\textwidth]{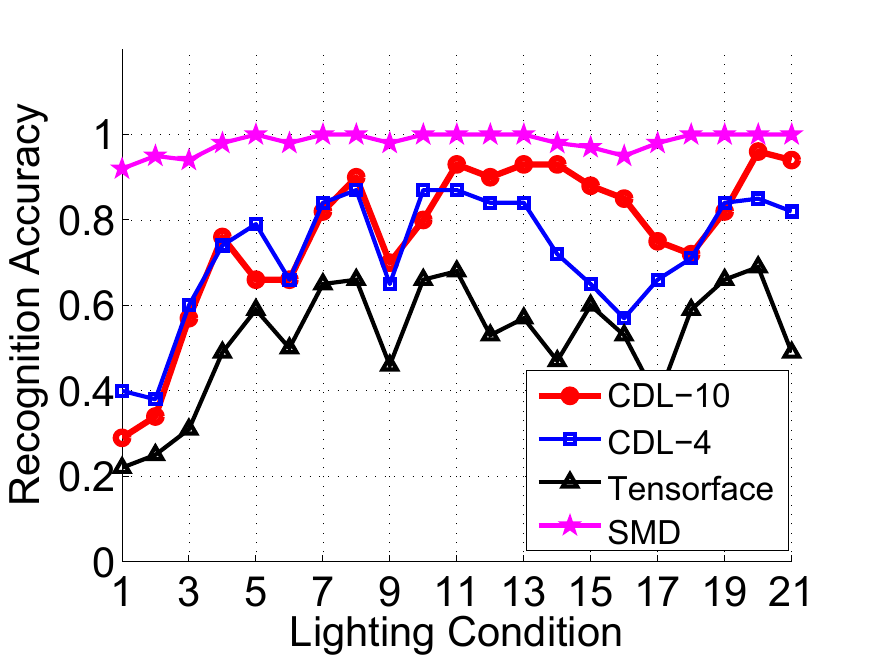} }
\\    \subfloat[Gallery: frontal. Probe: profile.] {\label{fig:acc-FP} \includegraphics[angle=0, height=0.18\textwidth, width=.33\textwidth]{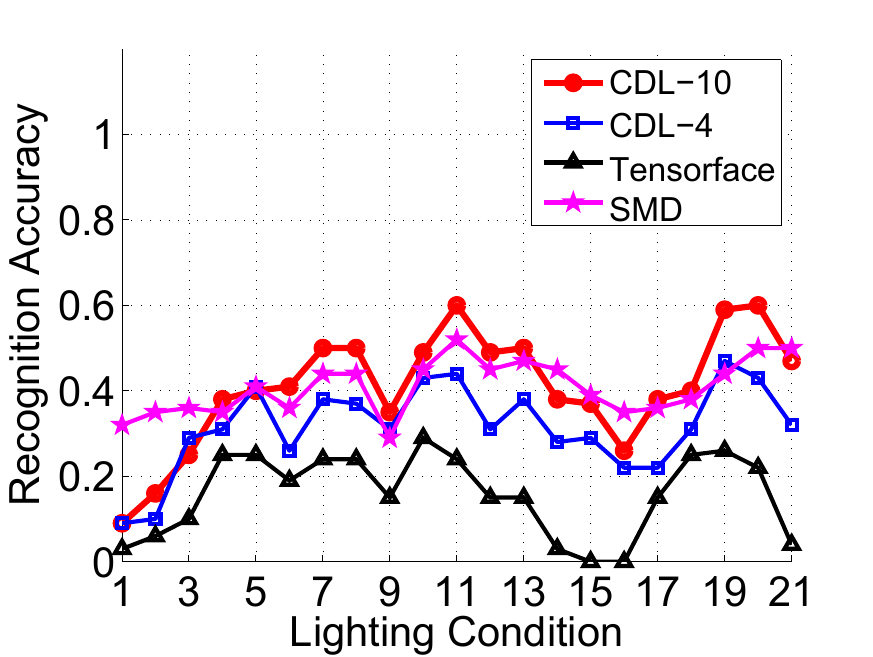} }
     \subfloat[Gallery: side. Probe: frontal.] {\label{fig:acc-SF} \includegraphics[angle=0, height=0.18\textwidth, width=.33\textwidth]{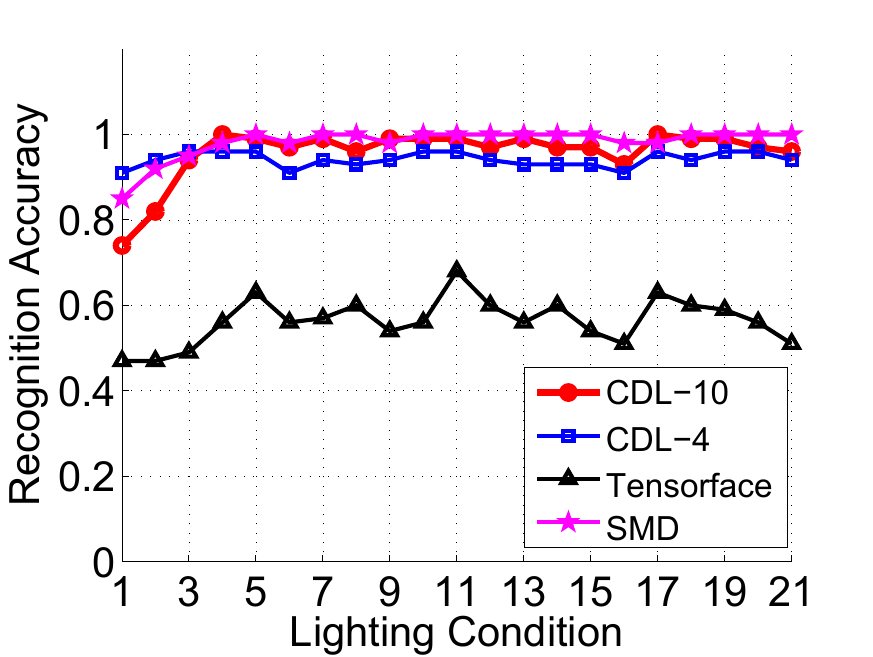} }
      \subfloat[Gallery: side. Probe: profile.] {\label{fig:acc-SP} \includegraphics[angle=0, height=0.18\textwidth, width=.33\textwidth]{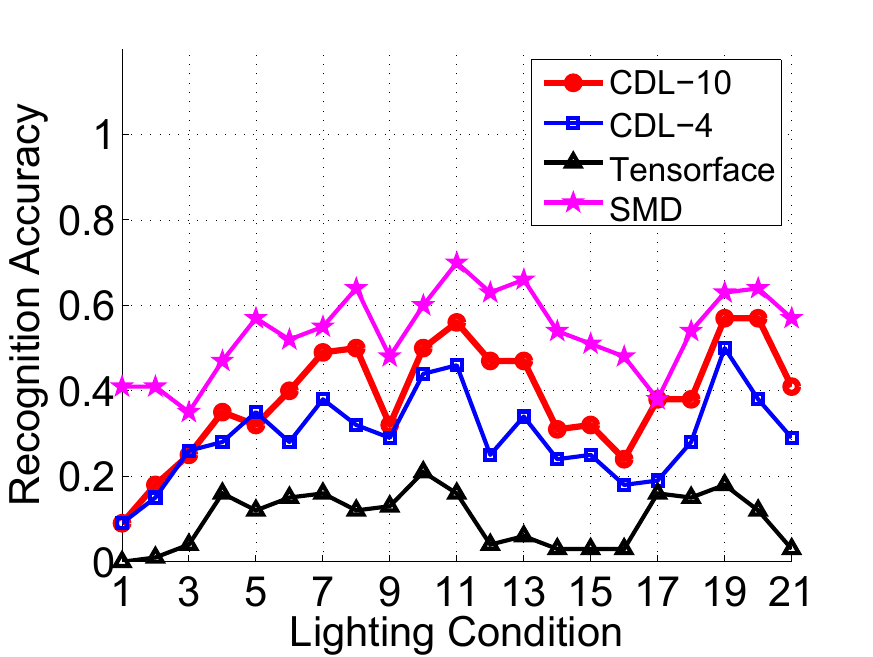} }
\caption{Face recognition under combined pose and illumination variations for the CMU PIE dataset. Given three testing poses, Frontal ($c27$), Side ($c05$), Profile ($c22$), we show the percentage of correct recognition for each disjoint pair of Gallery-Probe  poses. See Fig.~\ref{fig:pie} for poses and lighting conditions. Methods compared here include Tensorface \cite{tensorface1,tensorface2}, SMD \cite{smd} and our compositional dictionary learning (CDL) method . CDL-4 uses the dictionary $\mathbf{D}_{4}$ and CDL-10 uses $\mathbf{D}_{10}$. To the best of our knowledge, SMD reports the best recognition performance in such experimental setup.
4 out of 6 Gallery-Probe pose pairs, i.e., (a), (b), (d) and (e), our results are comparable to SMD.}
\label{fig:pie-acc}
\end{figure*}

Fig.~\ref{fig:pie-acc} shows the face recognition performance under combined pose and illumination variation for the CMU PIE dataset. To enable the comparison with \cite{smd}, we adopt the same challenging setup as described in \cite{smd}.
In this experiment, we classify 68 subjects in three poses, frontal ($c27$), side ($c05$), and profile ($c22$), under all 21 lighting conditions. We select one of the 3 poses as the gallery pose, and one of the remaining 2 poses as the probe pose,  for a total of 6 gallery-probe pose pairs. For each pose pair, the gallery is under the lighting condition $f11$ as specified in \cite{smd}, and the probe is under the illumination indicated in the table.
Methods compared here include Tensorface\cite{tensorface1,tensorface2},  SMD \cite{smd}, and the proposed method CDL. CDL-4 uses the dictionary $\mathbf{D}_{4}$ and CDL-10 uses $\mathbf{D}_{10}$. In both CDL-4 and CDL-10 setups, three testing poses $c27$, $c05$, and $c22$ are unknown to the training data. It is noted that, to the best of our knowledge, SMD reports the best recognition performance in such experimental setup.
As shown in Fig.~\ref{fig:pie-acc}, among 4 out of 6 Gallery-Probe pose pairs, the proposed CDL-10 is better or comparable to SMD.

The stereo matching distance method performs classification based on the stereo matching distance between each pair of gallery-probe images.
The stereo matching method can be seen as an example of a zero-shot method as no training is involved.
The stereo matching distance becomes more robust when the pose variation between such image pair decreases. However, the proposed CDL classifies faces based on subject codes extracted from each image alone. The robustness of the extracted subject codes only depends on the capability of the base dictionary to reconstruct such a face.
This explains why our CDL method significantly outperforms SMD for more challenging pose pairs, e.g., \emph{Profile-Frontal} pair with $62^o$ pose variation; but performs worse than SMD for easier pairs, e.g., \emph{Frontal-Side} with $16^o$ pose variation.

It can be observed in Fig.~\ref{fig:d68unknown} that an unknown pose can be approximated in terms of a set of observed poses. By representing three testing poses through four training poses in $\mathbf{D}_{4}$, instead of ten poses in $\mathbf{D}_{10}$, we obtain reasonable performance degradations but with 60\% less training data.

Though the Tensorface method shares a similar multilinear framework to CDL, as seen from Fig.~\ref{fig:pie-acc}, it only handles limited pose and illumination variations.

\subsubsection{Classifying Extended YaleB using $\mathbf{D}_{32}$}

\begin{figure} [ht]
\centering
\includegraphics[angle=0, height=0.1\textwidth, width=.5\textwidth]{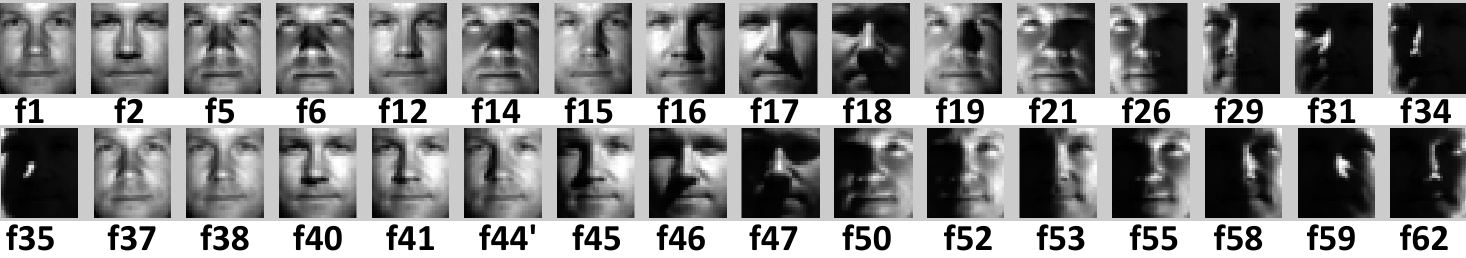}
\caption{Illumination variation in the Extended YaleB dataset.}
\label{fig:yaleb_light}
\end{figure}

We adopt a similar protocol as described in \cite{lcksvd}. In the Extended YaleB dataset, each of the 38 subjects is imaged under 64 lighting conditions. We split the dataset into two halves by randomly selecting 32 lighting conditions as training, and the other half for testing.
Fig.~\ref{fig:yaleb_light} shows the illumination variation in the testing data. When we learn $\mathbf{D}_{32}$ using Algorithm~\ref{algo:basedict},
we also obtain one unique domain-invariant sparse representation for each subject.
 During testing, we extract the subject codes from each testing face image and classify it based on the best match in unique sparse representation of each subject learned during training.
As shown in Table~\ref{tab:yalebacc}, the proposed CDL method outperforms other state-of-the-art sparse representation methods (The results for other compared methods are taken from \cite{lcksvd}).
When the extreme illumination conditions are included, we obtain an average recognition rate 98.91\%.
By excluding the extreme illumination condition $f35$, we obtain an average recognition rate 99.7\%.

\begin{table*}[ht]
\centering
\caption{Face recognition rate (\%) on the Extended YaleB face dataset across 32 different lighting conditions. By excluding the extreme illumination condition f35, we obtain an average recognition rate 99.7\%}
{
\begin{tabular}{|l|l|l|l|l|l|}
\hline
\textbf{CDL} & D-KSVD \cite{Zhang10}& LC-KSVD \cite{lcksvd} & K-SVD \cite{Elad_KSVD}& SRC \cite{Wright09}& LLC \cite{llc-cvpr10}\\
\hline
\hline
\textbf{98.91} & 94.10  & 95.00 & 93.1 & 97.20 & 90.7\\
\hline
\end{tabular}
}
\label{tab:yalebacc}
\end{table*}

\subsubsection{Comparisons with More Face Recognition Methods}

\begin{figure*} [t]
\centering
 \subfloat[Pose c02] {\label{fig:eccv12-c02} \includegraphics[angle=0, height=0.18\textwidth, width=.25\textwidth]{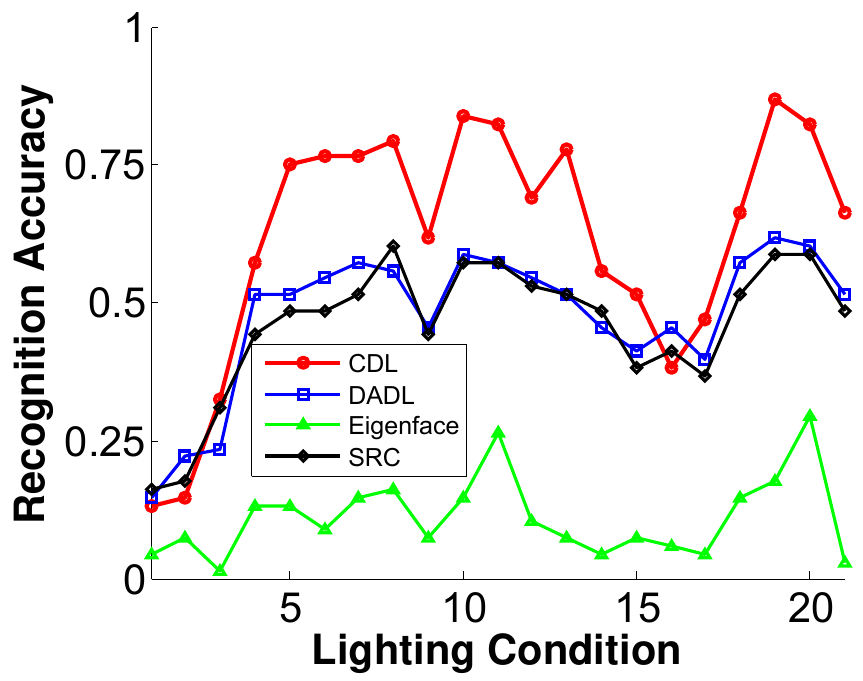} \hspace{0pt}}
  \subfloat[Pose c05] {\label{fig:eccv12-c05} \includegraphics[angle=0, height=0.18\textwidth, width=.25\textwidth]{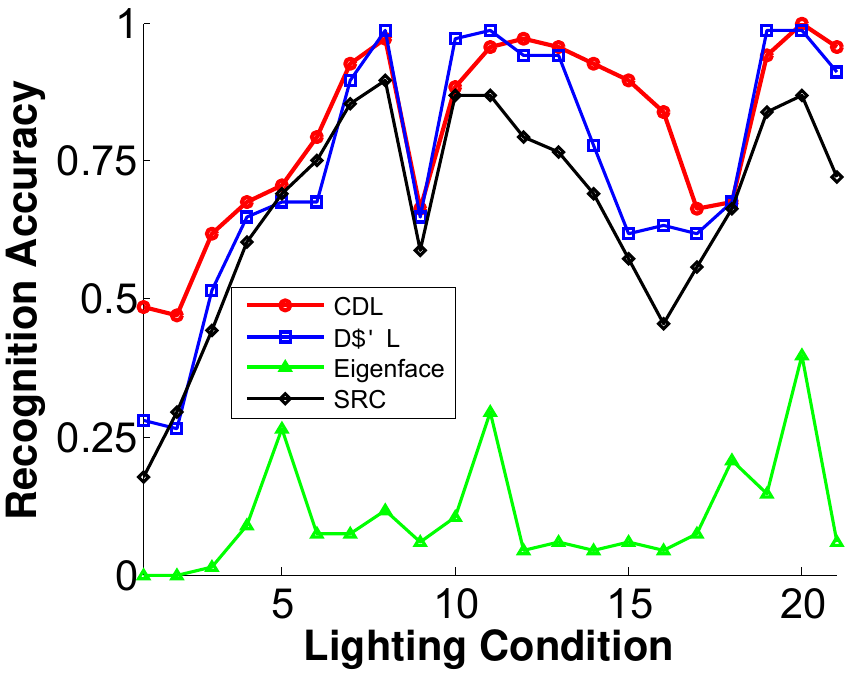}}
    \subfloat[Pose c29] {\label{fig:eccv12-c29} \includegraphics[angle=0, height=0.18\textwidth, width=.25\textwidth]{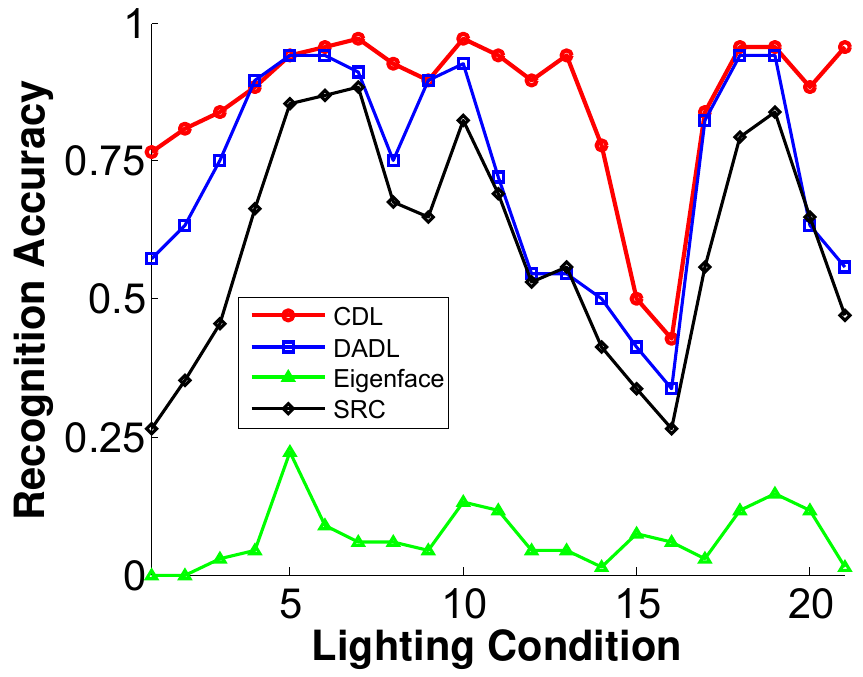}}
  \subfloat[Pose c14] {\label{fig:eccv12-c14} \includegraphics[angle=0, height=0.18\textwidth, width=.25\textwidth]{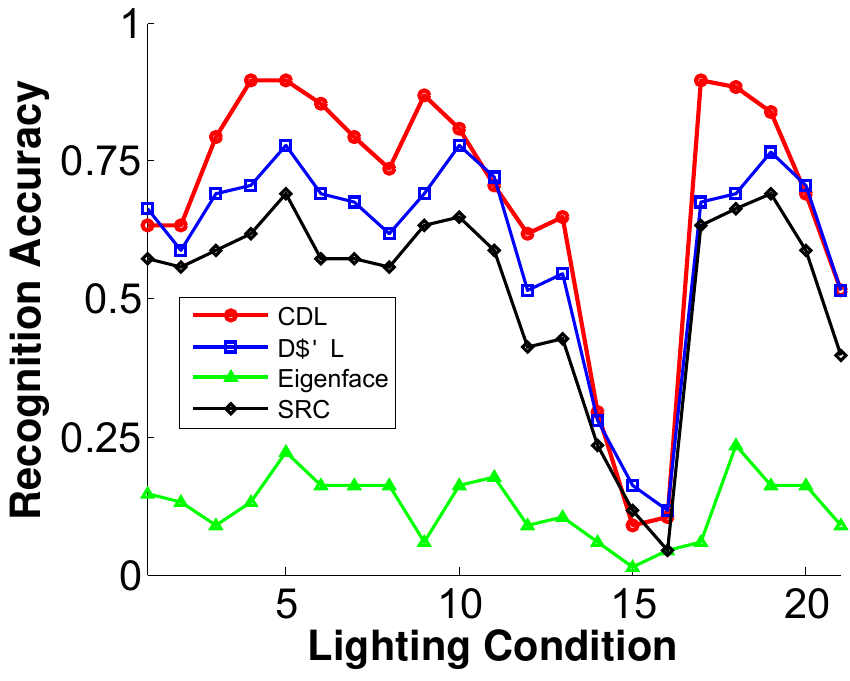}}
\caption{Face recognition accuracy on the CMU PIE dataset using the experimental protocol in \cite{dadl}.
The domain base dictionary is learned from five training poses $c11$, $c22$ $c27$, $c34$, and $c37$.
The classification accuracy is reported on 68 subjects 5712 face images in four testing poses $c02$, $c05$, $c14$, and $c29$ over 21 different lighting conditions.
The proposed method is denoted as CDL in color red. The proposed method significantly outperforms state-of-the-art methods DADL \cite{dadl} and SRC \cite{Wright09} for face recognition across domains.}
\label{fig:pieeccv}
\end{figure*}

In this section, we present comparisons with more state-of-the-art face recognition methods to further evaluate the effectiveness of our approach for face recognition across domains.
In \cite{dadl}, a different approach for realizing domain-adaptive face recognition is presented.
We adopt the same experimental conditions in \cite{dadl} by learning a domain base dictionary using five training poses $c11$, $c22$ $c27$, $c34$, and $c37$.
Fig.~\ref{fig:pieeccv} shows the classification accuracy on 68 subjects 5712 face images
 in four testing poses $c02$, $c05$, $c14$, and $c29$ over 21 different lighting conditions. The proposed method (color \emph{red}) significantly outperforms state-of-the-art methods DADL \cite{dadl} and SRC \cite{Wright09} for face recognition across domains.

We further compare with several techniques designed for illumination robust face representation, including Gradientfaces \cite{face-D}, LTV \cite{LTV}, SQI \cite{SQI},  and MSR \cite{MSR}.
Following the experiments described in \cite{face-D},  we use 68 subjects with 1428 frontal ($c27$) face images, each with 21 different illuminations for testing.  We use one image per subject as the reference images, the other images as the query images.
It is noted that some of the compared methods here are unsupervised, and the proposed method requires an additional base dictionary learning step. We adopt here the domain base dictionary $\mathbf{D}_{4}$ learned from four other training poses. As shown in Table~\ref{tab:pieacc2}, the proposed method outperforms compared methods for face recognition under varying illumination.

As discussed, in Table~\ref{tab:yalebacc},  we obtain one unique domain-invariant sparse representation for each subject during the domain base dictionary learning. During testing, we extract the subject codes from each testing face image and classify it based on the best match in unique sparse representation of each subject learned from the training data.
We now adopt a different experimental protocol on the extended YaleB dataset  as discussed in \cite{face-E}. We randomly select 5 images per subject from the training data as the reference and use the remaining images as the query images. The same base dictionary $\mathbf{D}_{32}$ is adopted. We obtain recognition accuracy $99.80\%$, which is comparable to $97.80\%$ reported in \cite{face-E}.

\begin{table*}[ht]
\centering
\caption{Face recognition rate (\%) on the PIE face dataset (pose $c27$) under varying illumination.}
{
\begin{tabular}{|l|l|l|l|l|}
\hline
\textbf{CDL} &  Gradientfaces \cite{face-D}& LTV \cite{LTV} & SQI \cite{SQI}& MSR \cite{MSR} \\
\hline
\hline
\textbf{99.93} & 99.83  & 86.85 & 77.94 & 62.07 \\
\hline
\end{tabular}
}
\label{tab:pieacc2}
\end{table*}

\begin{figure*} [ht]
\centering
 \subfloat[Frontal pose ($c27$).] {\label{fig:subcode1_front} \includegraphics[angle=0, height=0.16\textwidth, width=.3\textwidth]{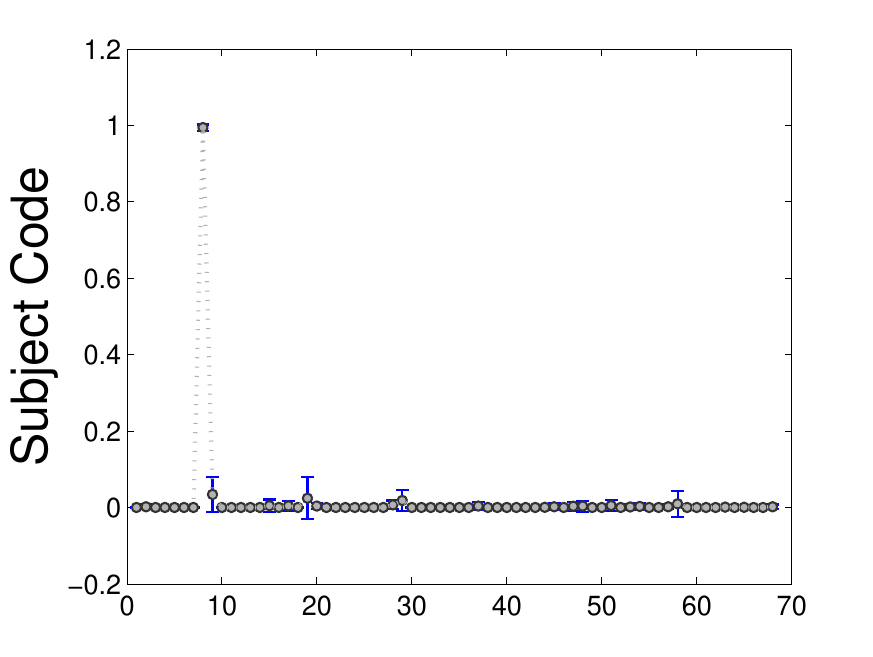} }
 \subfloat[Side pose ($c05$).] {\label{fig:subcode1_side} \includegraphics[angle=0, height=0.16\textwidth, width=.3\textwidth]{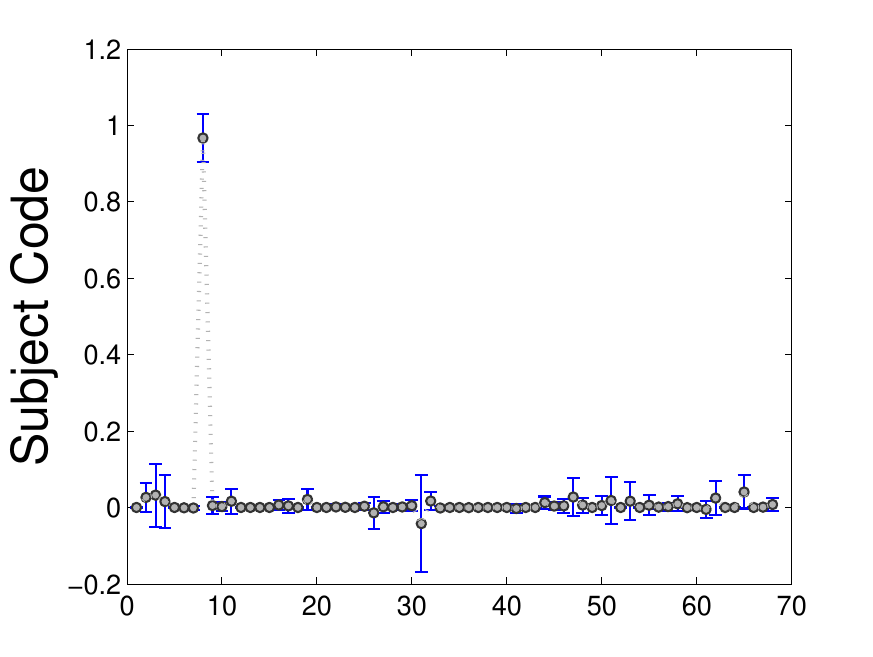}}
  \subfloat[Profile pose ($c22$).] {\label{fig:subcode1_profile} \includegraphics[angle=0, height=0.16\textwidth, width=.3\textwidth]{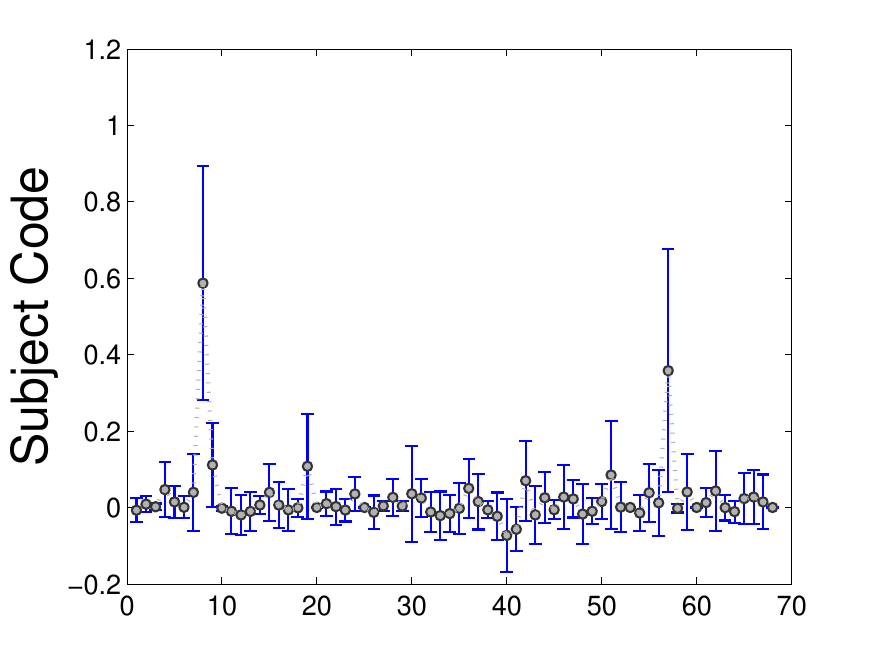}} \\
 \subfloat[Frontal pose ($c27$).] {\label{fig:TF_subcode1_front} \includegraphics[angle=0, height=0.16\textwidth, width=.3\textwidth]{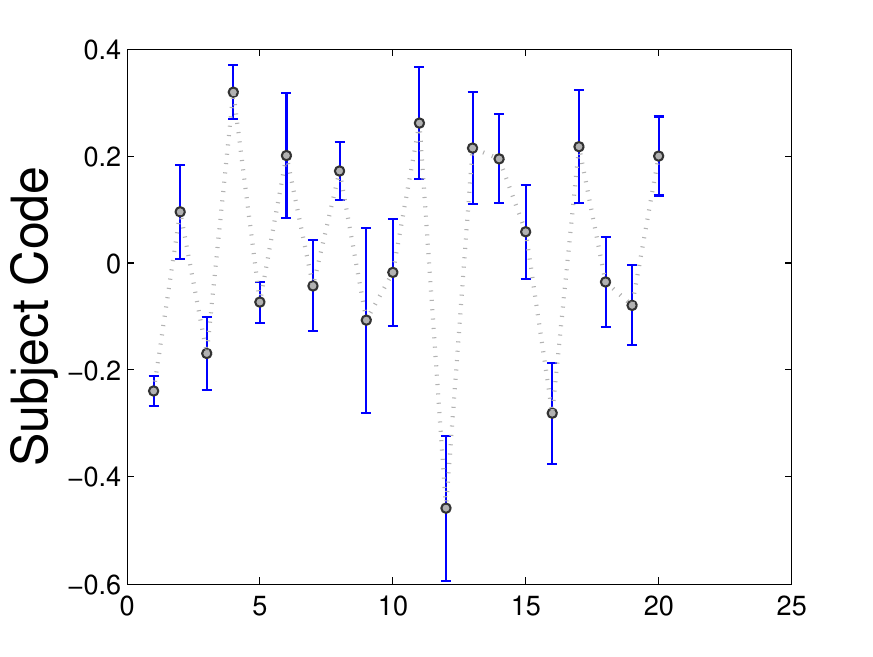} }
 \subfloat[Side pose ($c05$).] {\label{fig:TF_subcode1_side} \includegraphics[angle=0, height=0.16\textwidth, width=.3\textwidth]{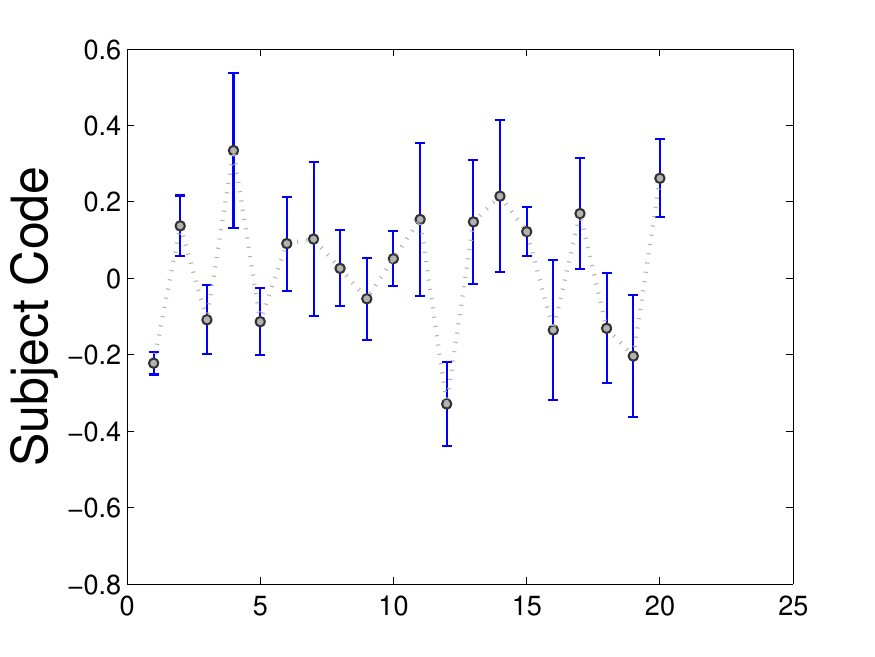}}
  \subfloat[Profile pose ($c22$).] {\label{fig:TF_subcode1_profile} \includegraphics[angle=0, height=0.16\textwidth, width=.3\textwidth]{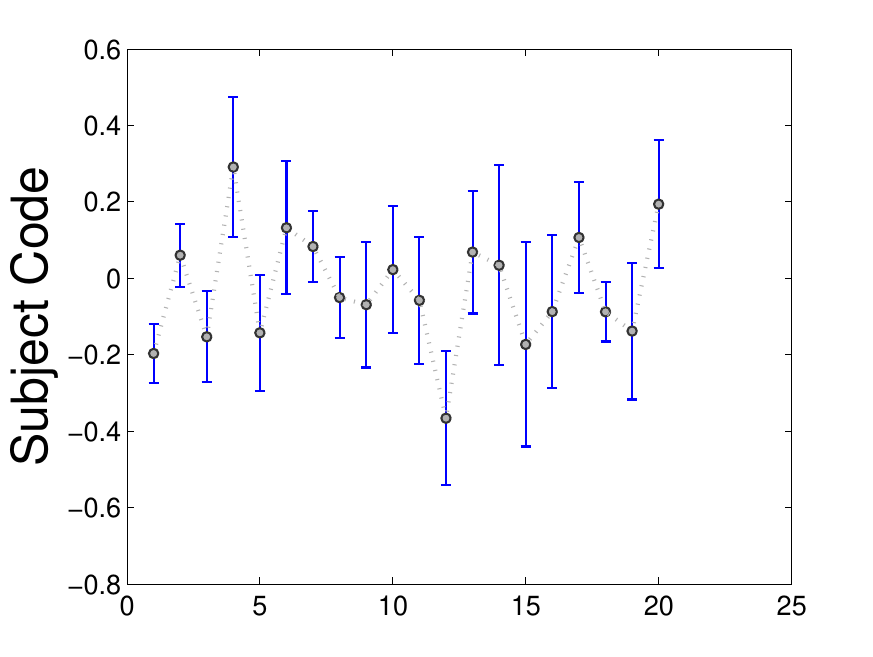}}
\caption{Mean subject code of subject $s1$ over 21 illumination conditions in each of the three testing poses, and standard error of the mean code. (a),(b),(c) are generated using CDL with the base dictionary $\mathbf{D_{10}}$. (d),(e),(f) are generated using Tensorfaces.}
\label{fig:subcode1}
\end{figure*}

\begin{figure*} [ht]
\centering
 \subfloat[Frontal pose ($c27$).] {\label{fig:subcode2_front} \includegraphics[angle=0, height=0.16\textwidth, width=.3\textwidth]{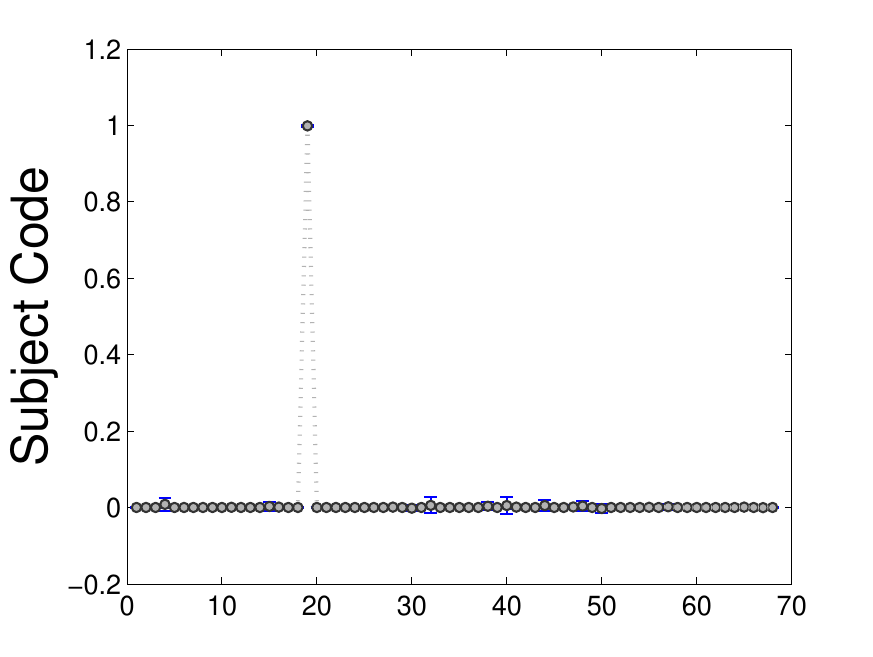} }
 \subfloat[Side pose ($c05$).] {\label{fig:subcode2_side} \includegraphics[angle=0, height=0.16\textwidth, width=.3\textwidth]{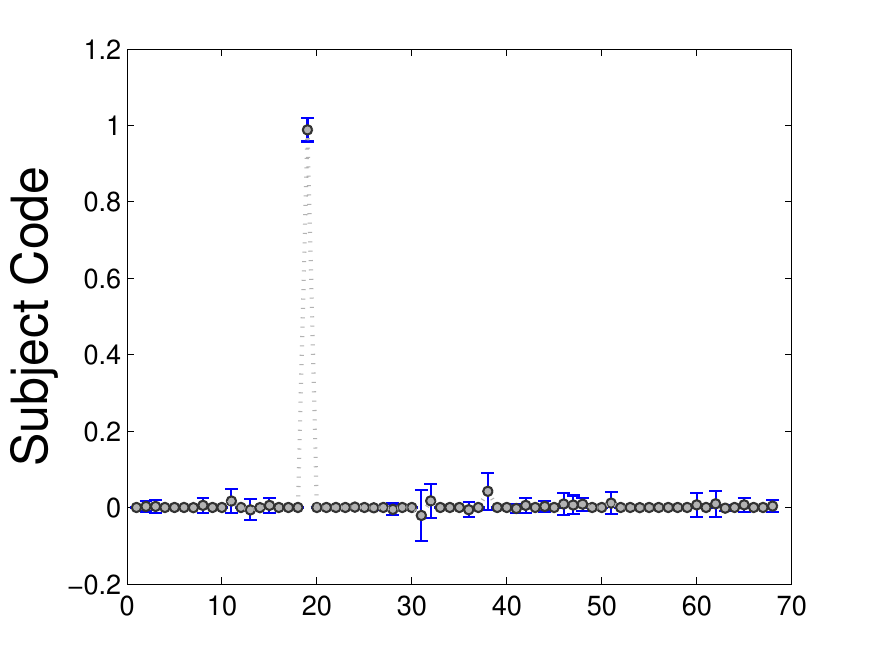}}
  \subfloat[Profile pose ($c22$).] {\label{fig:subcode2_profile} \includegraphics[angle=0, height=0.16\textwidth, width=.3\textwidth]{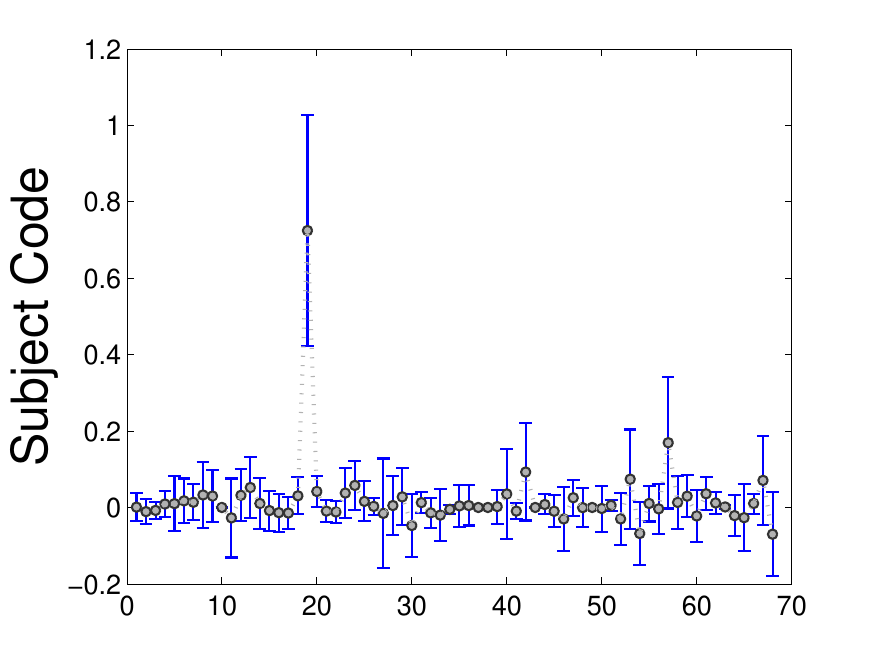}} \\
 \subfloat[Frontal pose ($c27$).] {\label{fig:TF_subcode2_front} \includegraphics[angle=0, height=0.16\textwidth, width=.3\textwidth]{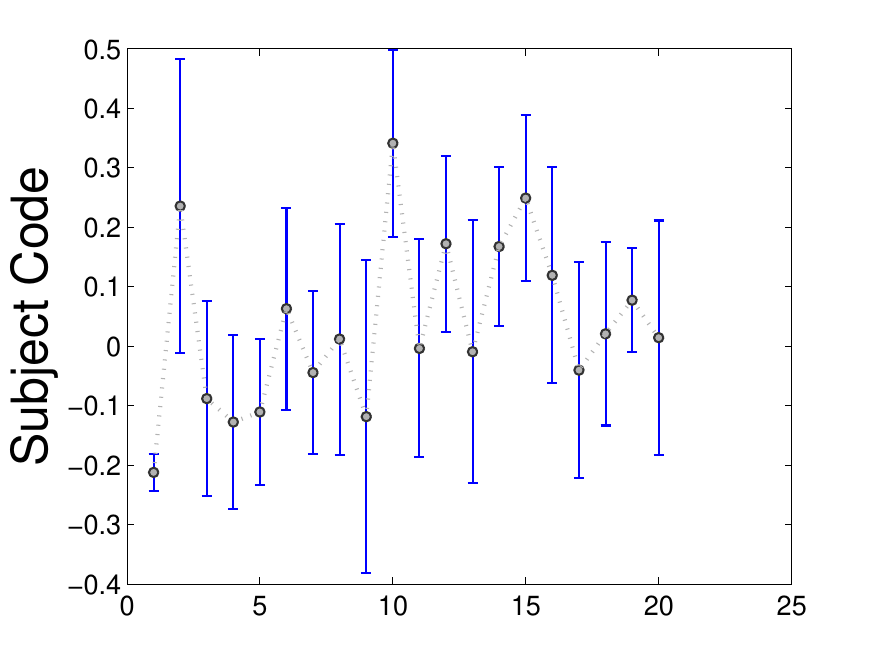} }
 \subfloat[Side pose ($c05$).] {\label{fig:TF_subcode2_side} \includegraphics[angle=0, height=0.16\textwidth, width=.3\textwidth]{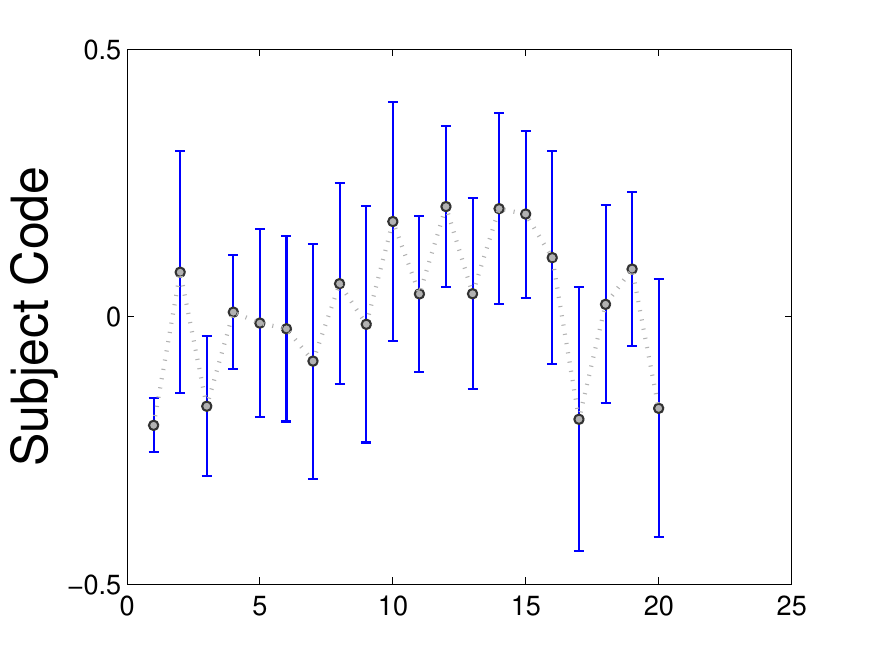}}
  \subfloat[Profile pose ($c22$).] {\label{fig:TF_subcode2_profile} \includegraphics[angle=0, height=0.16\textwidth, width=.3\textwidth]{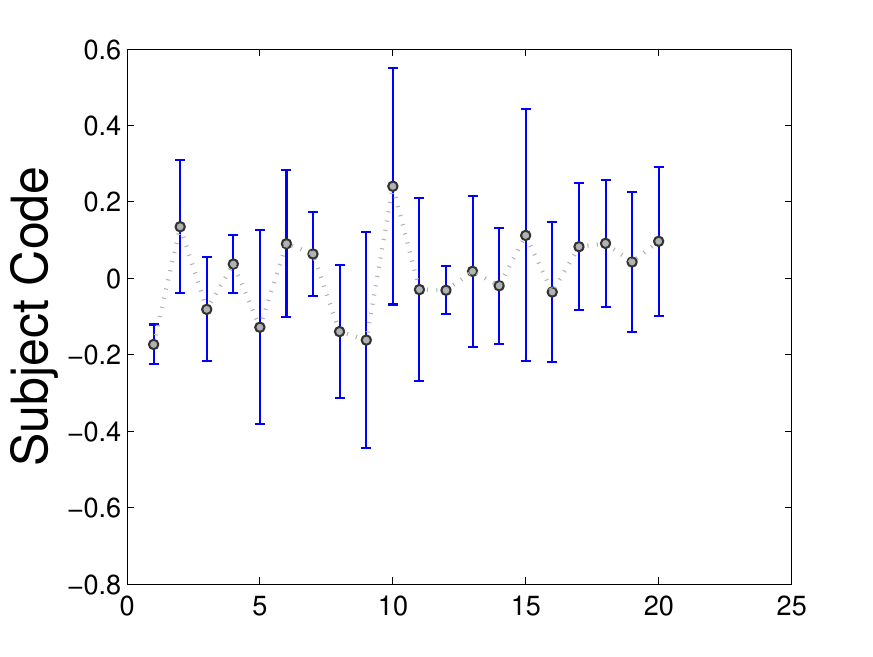}}
\caption{Mean subject code of subject $s2$ over 21 illumination conditions in each of the three testing poses, and standard error of the mean code. (a),(b),(c) are generated using CDL with the base dictionary $\mathbf{D_{10}}$. (d),(e),(f) are generated using Tensorfaces.}
\label{fig:subcode2}
\end{figure*}

\subsection{Mean Code and Error Analysis}
\label{errorcode}

As discussed in Sec.~\ref{sec:tensor}, the Tensorface method shares a similar multilinear framework to the proposed CDL method. However, we showed through the above experiments
that the proposed method based on sparse decomposition significantly outperforms the $N$-mode SVD decomposition for face recognition across pose and illumination. In this section, we analyze in more detail the behaviors of the proposed CDL and Tensorfaces,
by comparing subject and domain codes extracted from a face image using these two methods.

For the experiments in this section, we adopt the base dictionary $\mathbf{D}_{10}$ for CDL, and the same training data and sparsity values of $\mathbf{D}_{10}$ for Tensorfaces to learn the core tensor and the associated mode matrices. The same testing data is used for both methods, i.e., 68 subjects in the PIE dataset under 21 illumination conditions in the $c27$ (frontal), $c05$ (side) and $c22$ (profile) poses, which are three unseen poses not present in the training data.

Fig.~\ref{fig:subcode1} and Fig.~\ref{fig:subcode2} shows the mean subject codes of subject $s1$ and $s2$ over 21 illumination conditions in each of the three testing poses, and the associated standard errors.  In each of the two figures, we compare the first row, the subject codes from CDL, with the second row, the subject codes from Tensorfaces. We can easily notice the following: first, the subject codes extracted using CDL are more sparse; second,
CDL subject codes are more consistent across pose; third, CDL subject codes are more consistent across illumination, which is indicated by the smaller standard errors. By comparing Fig.~\ref{fig:subcode1} with Fig.~\ref{fig:subcode2}, we also observe that the CDL subject codes are more discriminative.
Table~\ref{tab:sqrtvar} further shows the square root of the pooled variances of subject codes for all 68 subjects over 21 illumination conditions in each of the three testing poses. The significantly smaller variance values obtained using CDL indicate the more consistent sparse representation of subjects decomposed from face images.
Therefore, face recognition using CDL subject codes  significantly outperforms recognition using Tensorfaces subject codes.

\begin{table*}[ht]
\centering
\caption{The square root of the pooled variances of subject codes for 68 subjects over 21 illumination conditions in each of the three testing poses.}
{
\begin{tabular}{|l|l|l|l|}
\hline
 & Frontal pose (c27) & Side pose (c05) & Profile pose (c22) \\
\hline
\hline
CDL & 0.0351  &  0.0590  &  0.0879 \\
\hline
Tensorfaces & 0.1479  &  0.1758  &  0.1814 \\
\hline
\end{tabular}
}
\label{tab:sqrtvar}
\end{table*}

\subsection{Pose and Illumination Estimation}
\label{domainEst}

\begin{figure} [ht]
\centering
 \subfloat[$\mathbf{D}_{10}$ ] {\label{fig:lightacc34} \includegraphics[angle=0, height=0.16\textwidth, width=0.15\textwidth]{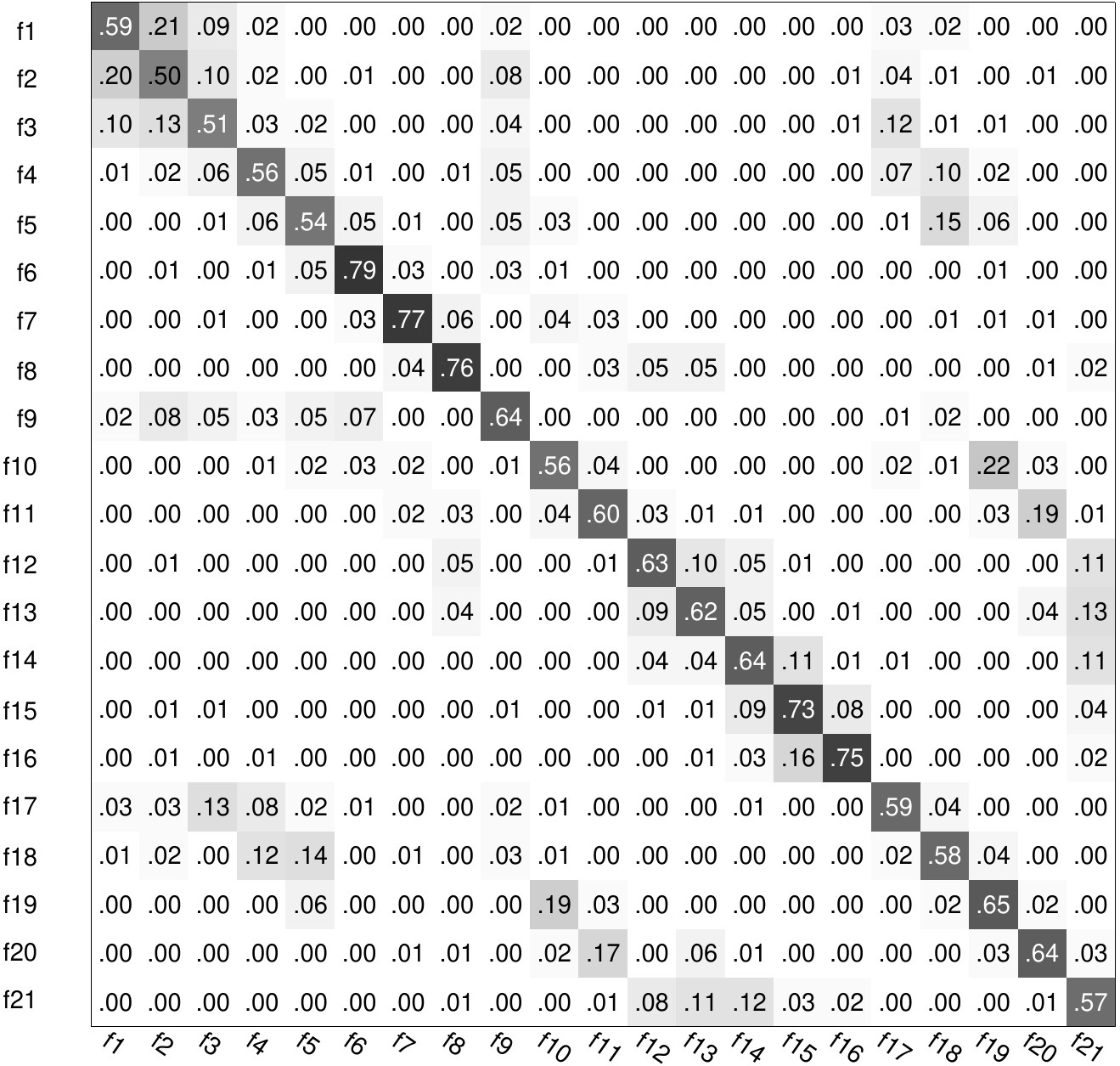} \hspace{5pt}}
  \subfloat[$\mathbf{D}_{4}$] {\label{fig:lightacc34} \includegraphics[angle=0, height=0.16\textwidth, width=0.15\textwidth]{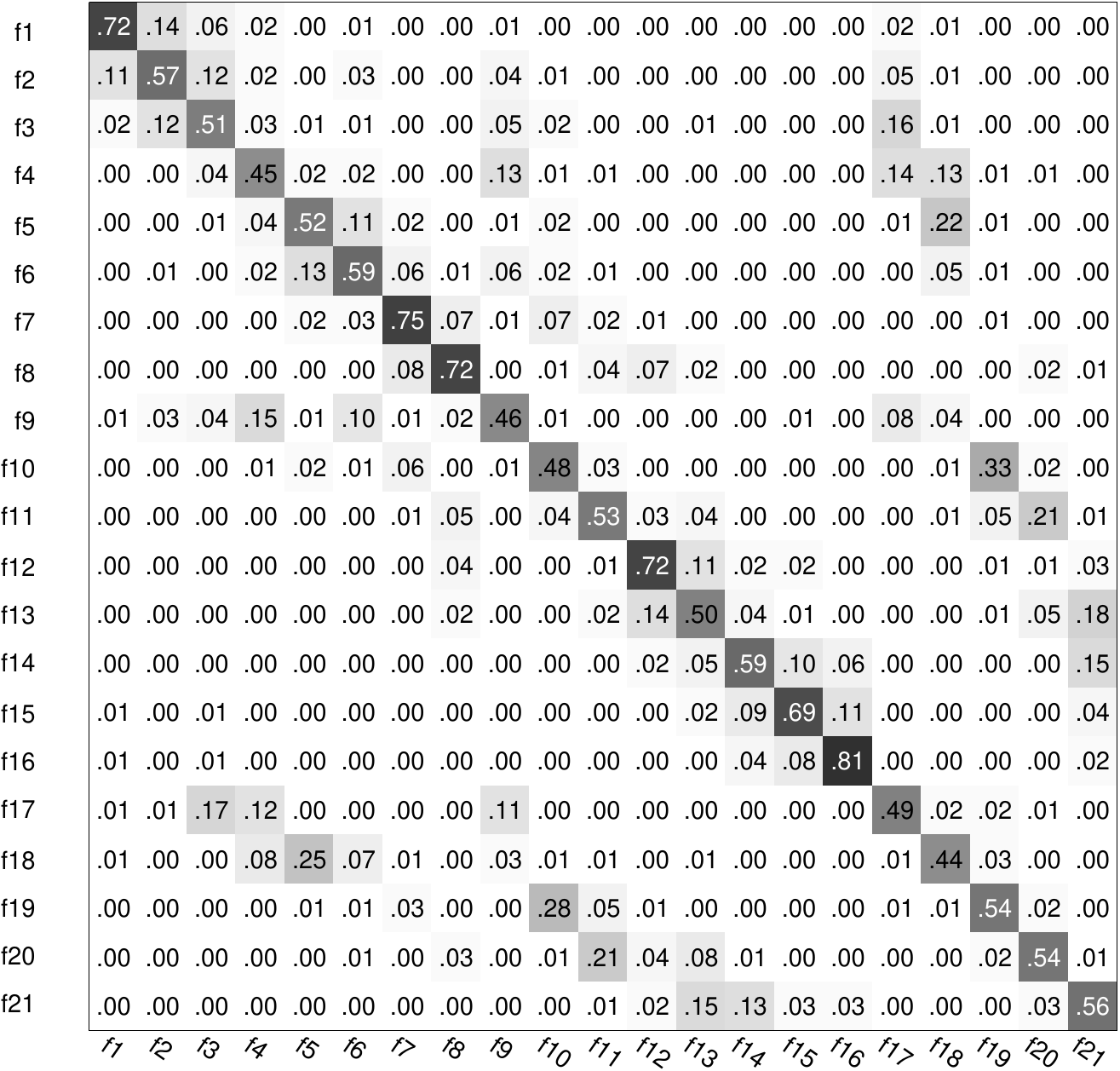}}
   \subfloat[ Tensorfaces ] {\label{fig:lightTF_sparse} \includegraphics[angle=0, height=0.16\textwidth, width=0.15\textwidth]{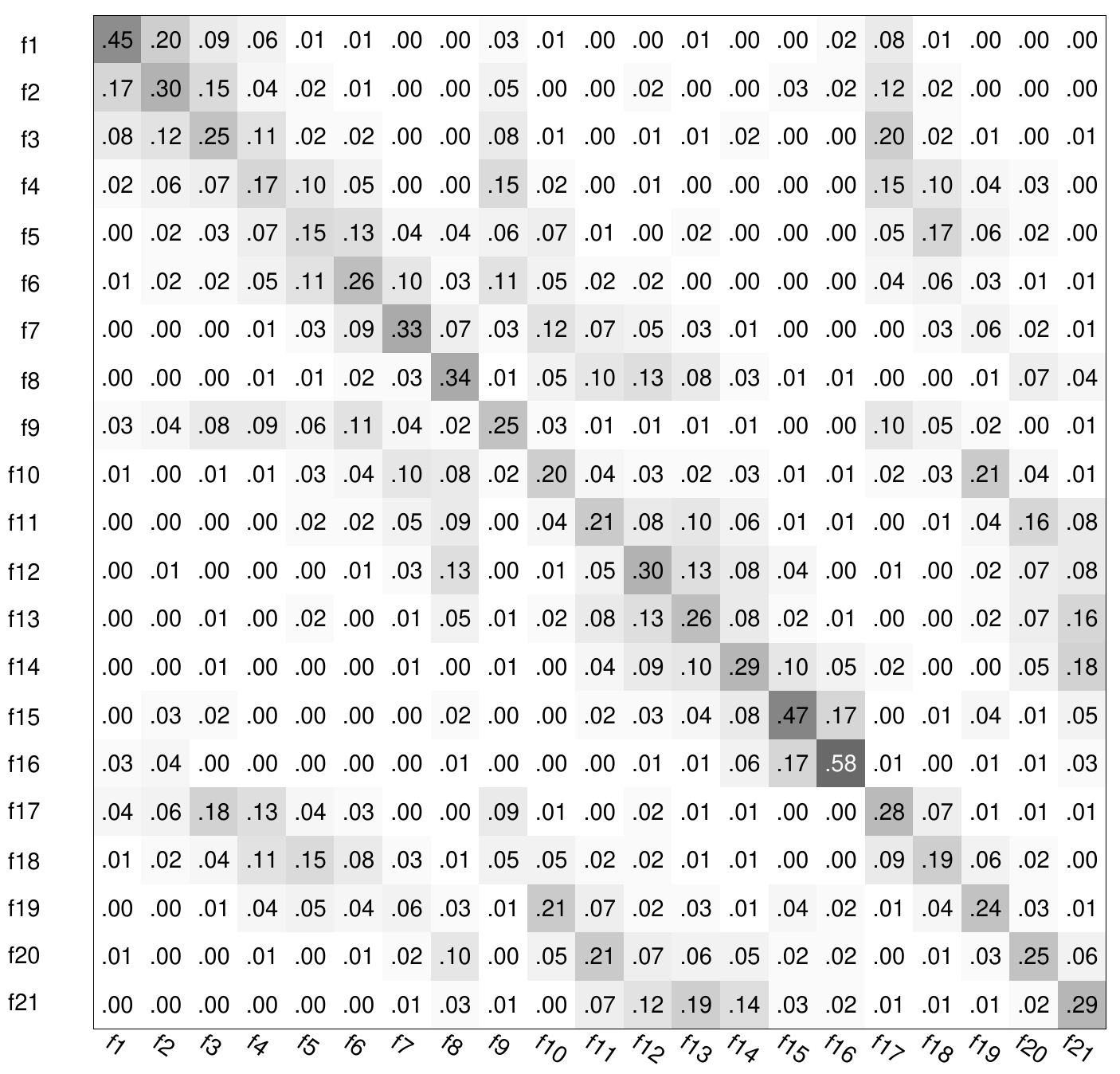}}
\\   \subfloat[$\mathbf{D}_{10}$] {\label{fig:poseacc34} \includegraphics[angle=0, height=0.16\textwidth, width=0.15\textwidth]{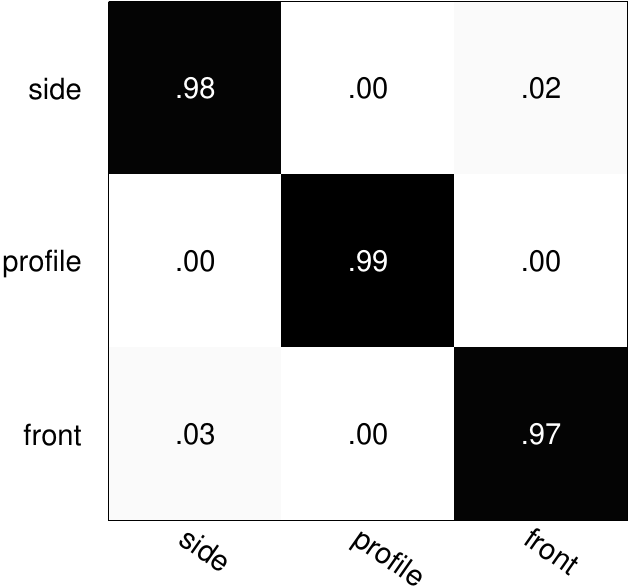} \hspace{5pt}}
  \subfloat[$\mathbf{D}_{4}$] {\label{fig:poseacc34} \includegraphics[angle=0, height=0.16\textwidth, width=0.15\textwidth]{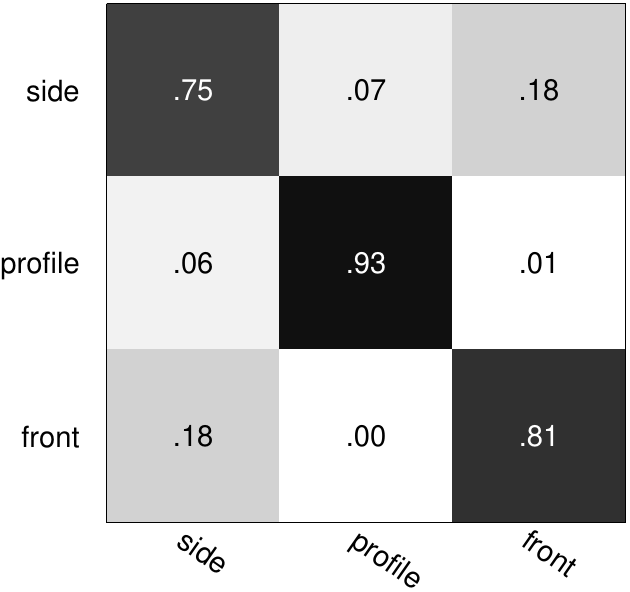}}
    \subfloat[Tensorfaces] {\label{fig:poseaccTF} \includegraphics[angle=0, height=0.16\textwidth, width=0.15\textwidth]{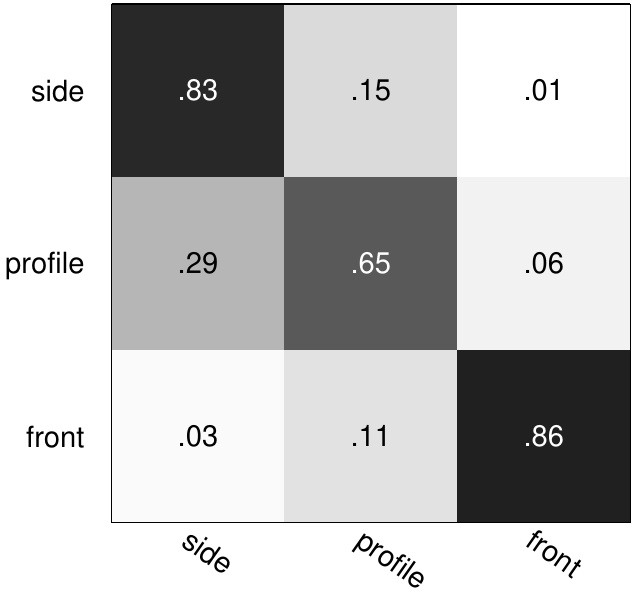}}
\caption{Illumination (a-c) and pose (d-f) estimation on the CMU PIE dataset using base dictionaries $\mathbf{D}_{4}$ and $\mathbf{D}_{10}$. Average accuracy: (a) 0.63, (b) 0.58, (c) 0.28, (d) 0.98, (e) 0.83, (f) 0.78. The proposed CDL method exhibits significantly better domain estimation accuracy than the Tensorfaces method. }
\label{fig:estacc68}
\end{figure}

In Section~\ref{sec:facerec}, we report the results of experiments over subject codes using base dictionaries $\mathbf{D}_{10}$ and $\mathbf{D}_{4}$.
While generating subject codes, we simultaneously obtain pose codes and illumination codes.
Such pose and illumination codes can be used for pose and illumination estimation.
In Fig.~\ref{fig:estacc68},  we show the pose and illumination estimation performance on the PIE dataset using the pose and illumination sparse codes through both CDL and Tensorfaces.
 The proposed CDL method exhibits significantly better domain estimation accuracy than the Tensorfaces method.
By examining Fig.~\ref{fig:estacc68}, it can be noticed that the most confusing illumination pairs in CDL, e.g., $(f05, f18)$, $(f10, f19)$ and $(f11, f20)$ are very visually similar based on Fig.~\ref{fig:pie}.

\section{Conclusion}
\label{sec:conclusion}
We presented an approach to learn domain adaptive dictionaries for face recognition across pose and illumination domain shifts. With a learned domain base dictionary, an unknown face image is decomposed into subject codes, pose codes and illumination codes. Subject codes are consistent across domains, and enable pose and illumination insensitive face recognition. Pose and illumination codes can be used to estimate the pose and lighting condition of the face.
We  plan to evaluate the proposed framework in representing 3D faces.  A face image captured by a RGB-D camera provides projected 2D images at various poses. Together with synthesized light sources, we can construct the proposed domain base dictionary; and the learned dictionary can then be used to decompose any given 2D face image for domain-invariant subject representation.
We will also experiment the proposed method as a novel way to synthesize more training samples from unseen pose and illumination conditions.

\section*{Acknowledgments}
This research was partially supported by a MURI from the Office of Naval
research under the Grant N00014-10-1-0934..

{
\bibliographystyle{IEEEtran}
\bibliography{comdic}
}

\newcommand{\biospace}{\vspace{-4em}}
\biospace
\begin{IEEEbiography}[{\includegraphics[width=1in,height=1.25in,clip,keepaspectratio]{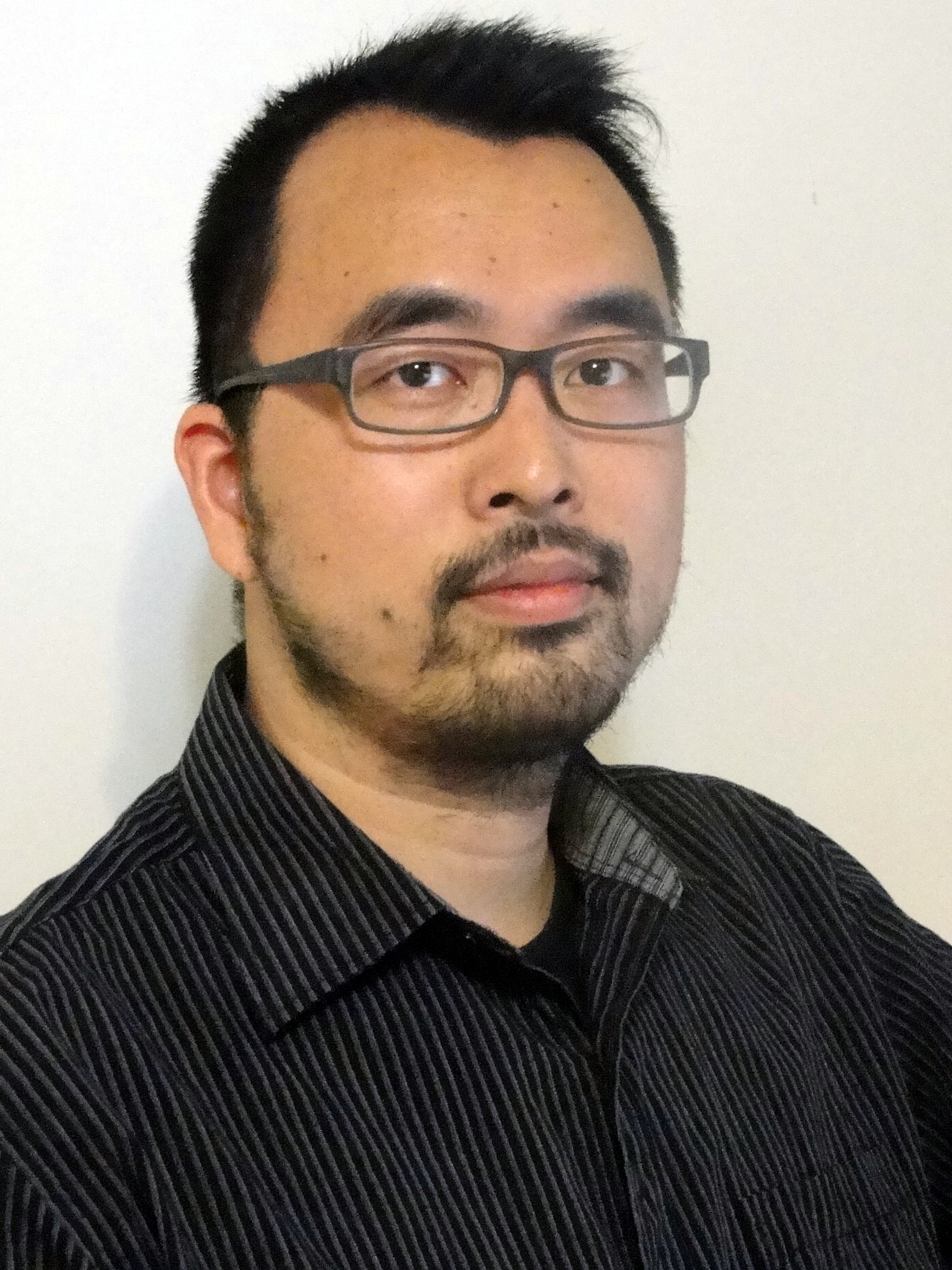}}]
{Qiang Qiu} received his Bachelor's degree with first class honors in Computer Science in 2001, and his Master's degree in Computer Science in 2002, from National University of Singapore. He received his Ph.D. degree in Computer Science in 2012 from  University of Maryland, College Park. During 2002-2007, he was a
  Senior Research Engineer at Institute for Infocomm Research, Singapore.  He is currently a Postdoctoral Associate at the Department of Electrical and Computer Engineering, Duke University. His research interests include computer vision and machine learning, specifically on face recognition, human activity recognition, image classification, and sparse representation.
\end{IEEEbiography}

\biospace
\begin{IEEEbiography}[{\includegraphics[width=1in,height=1.25in,clip,keepaspectratio]{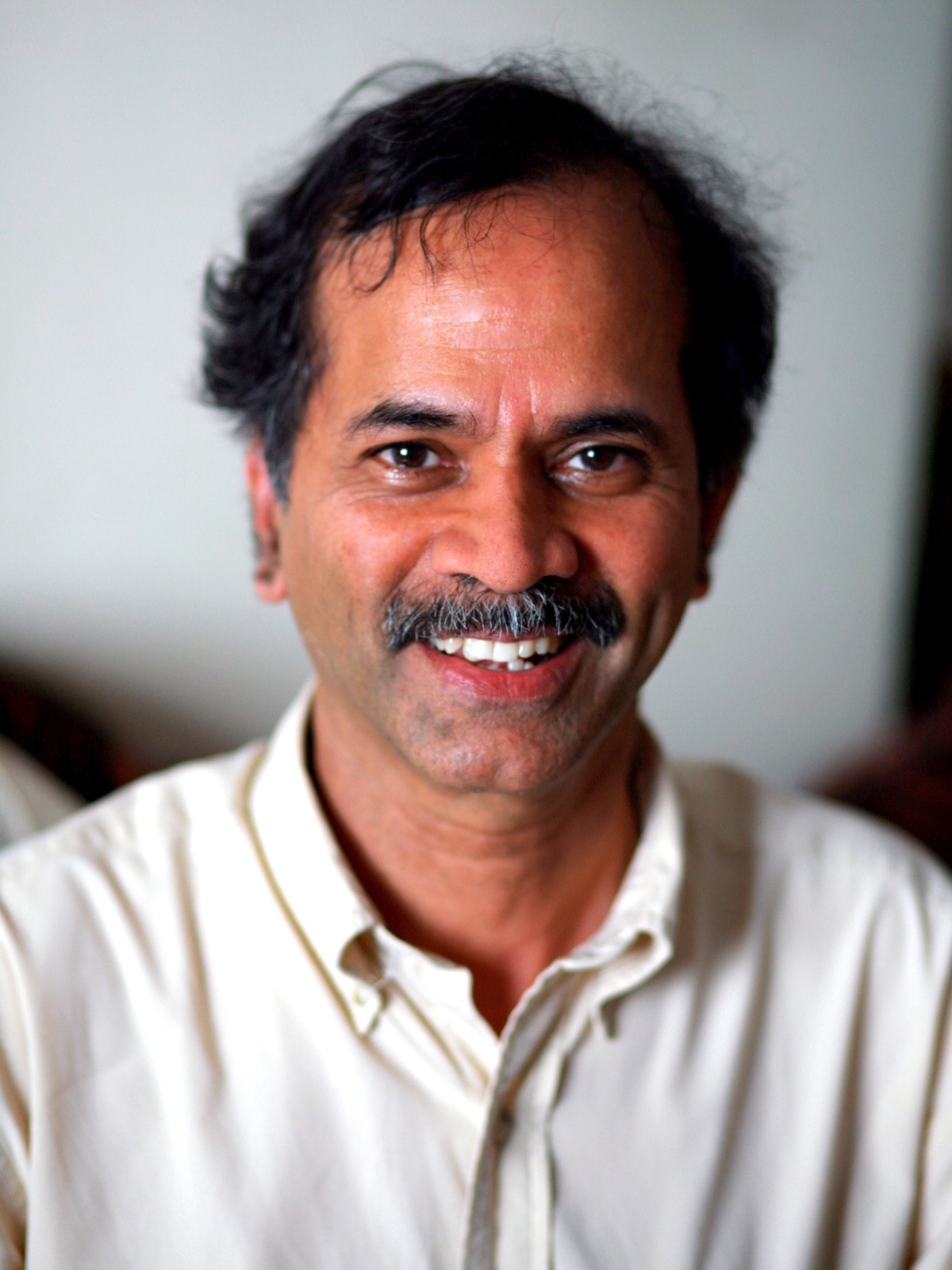}}]
{Rama Chellappa} received the B.E. (Hons.) degree in Electronics and Communication Engineering from the University of Madras, India and the M.E. (with Distinction) degree from the Indian Institute of Science, Bangalore, India. He received the M.S.E.E. and Ph.D. Degrees in Electrical Engineering from Purdue University, West Lafayette, IN. During 1981-1991, he was a faculty member in the department of EE-Systems at University of Southern California (USC). Since 1991, he has been a Professor of Electrical and Computer Engineering (ECE) and an affiliate Professor of Computer Science at University of Maryland (UMD), College Park.  He is also affiliated with the Center for Automation Research and the Institute for Advanced Computer Studies (Permanent Member) and is serving as the Chair of the ECE department.  In 2005, he was named a Minta Martin Professor of Engineering. His current research interests span many areas in image processing, computer vision and pattern recognition. Prof. Chellappa has received several awards including an NSF Presidential Young Investigator Award, four IBM Faculty Development Awards, two paper awards and the K.S. Fu Prize from the International Association of Pattern Recognition (IAPR). He is a recipient of the Society, Technical Achievement and Meritorious Service Awards from the IEEE Signal Processing Society. He also received the Technical Achievement and Meritorious Service Awards from the IEEE Computer Society. He is a recipient of Excellence in teaching award from the School of Engineering at USC. At UMD, he received college and university level recognitions for research, teaching, innovation and mentoring undergraduate students. In 2010, he was recognized as an Outstanding ECE by Purdue University. Prof. Chellappa served as the Editor-in-Chief of IEEE Transactions on Pattern Analysis and Machine Intelligence and as the General and Technical Program Chair/Co-Chair for several IEEE international and national conferences and workshops. He is a Golden Core Member of the IEEE Computer Society, served as a Distinguished Lecturer of the IEEE Signal Processing Society and as the President of IEEE Biometrics Council. He is a Fellow of IEEE, IAPR, OSA, AAAS, ACM and AAAI and holds four patents.
\end{IEEEbiography}

\end{document}